%% file: iclr2026_conference.tex
\documentclass{article} %
\usepackage{iclr2026_conference,times}

\input{math_commands.tex}

\usepackage{hyperref}
\usepackage{url}
\usepackage{xspace}
\usepackage{wrapfig}
\usepackage{algorithm}
\usepackage{algpseudocode}
\usepackage{amsmath}
\usepackage{xcolor}

\usepackage{graphicx}
\usepackage{booktabs}
\usepackage{colortbl}
\usepackage{multirow}
\usepackage{makecell}
\usepackage{cleveref}
\usepackage{bm}
\usepackage{pifont}

\usepackage[T1]{fontenc}   %
\usepackage{seqsplit}      %

\newcommand{\textttsplit}[1]{\texttt{\seqsplit{#1}}}

\usepackage{siunitx}        %
\newcommand{\bestscore}[1]{\textbf{\num{#1}}}
\newcommand{\secondscore}[1]{\underline{\num{#1}}}

\usepackage{etoc}

\title{You only need 4 extra tokens: \\
Synergistic Test-time Adaptation for LLMs}

\author{\begin{tabular}{@{}l@{}}
  Yijie Xu\textsuperscript{1},
  Huizai Yao\textsuperscript{1},
  Zhiyu Guo\textsuperscript{1},
  Pengteng Li\textsuperscript{1},\\
  Aiwei Liu\textsuperscript{3},
  Xuming Hu\textsuperscript{1,2},
  Weiyu Guo\textsuperscript{1,2,}\footnotemark[1],
  Hui Xiong\textsuperscript{1,2,}\thanks{Corresponding Authors.}
  \end{tabular}\\[0.5em]
  \textsuperscript{1}The Hong Kong University of Science and Technology (Guangzhou)\\
  \textsuperscript{2}The Hong Kong University of Science and Technology\\
  \textsuperscript{3}Tsinghua University\\[-0.1em]
  \texttt{\href{mailto:yxu409@connect.hkust-gz.edu.cn}{yxu409@connect.hkust-gz.edu.cn}}
}

\newcommand{\method}{\textsc{SyTTA}\xspace}

\newcommand{\modecache}{\textit{Static-Ref}\xspace}
\newcommand{\modewocache}{\textit{Dynamic-Ref}\xspace}
\newcommand{\rougelsum}{ROUGE-L$_\text{sum}$\xspace}

\renewcommand{\algorithmiccomment}[1]{\hfill\textcolor{gray}{\scriptsize\texttt{// #1}}}

\iclrfinalcopy %
\begin{document}

\maketitle

\input{tex/00_abstract}

\input{tex/01_intro}

\input{tex/02_related_work}

\input{tex/03_problem_setup}

\input{tex/04_method}

\input{tex/05_experiments}

\input{tex/06_findings}

\input{tex/07_conclusion}

\clearpage

\bibliography{iclr2026_conference}
\bibliographystyle{iclr2026_conference}

\clearpage

\input{tex/99_appendix}

\end{document}

%% file: math_commands.tex
\usepackage{amsmath,amsfonts,bm}

\def\eqref#1{equation~\ref{#1}}

\def\1{\bm{1}}

\DeclareMathAlphabet{\mathsfit}{\encodingdefault}{\sfdefault}{m}{sl}
\SetMathAlphabet{\mathsfit}{bold}{\encodingdefault}{\sfdefault}{bx}{n}

%% file: tex/00_abstract.tex
\vspace{-0.75em}
\begin{abstract}
\vspace{-0.25em}
Large language models (LLMs) are increasingly deployed in specialized domains such as finance, medicine, and agriculture, where they face significant distribution shifts from their training data. Domain-specific fine-tuning can mitigate this challenge but relies on high-quality labeled data that is expensive and slow to collect in expertise-limited settings.
We study label-free test-time adaptation for language models and present \method, an inference-time framework that adapts models on-the-fly without additional supervision. \method couples two complementary uncertainty signals that arise under distribution shift: input-side perplexity, indicating mismatch with domain-specific terminology and patterns, and output-side predictive entropy, indicating diffuse and unstable token probabilities during generation. %
Across diverse model architectures and domain-specific benchmarks, \method delivers consistent gains. Notably, on agricultural question answering, \method improves \rougelsum by over 120\% on \textsc{Qwen-2.5-7B} with only 4 extra tokens per query. These results show that effective test-time adaptation for language models is achievable without labeled examples, supporting deployment in label-scarce domains. 
The code will be made available upon acceptance.

\end{abstract}

%% file: tex/01_intro.tex
\vspace{-0.75em}
\section{Introduction}
\vspace{-0.5em}
\label{sec:intro}

Large language models (LLMs) have strong capabilities in reasoning, code generation, and language understanding, and they are being deployed in specialized domains or scenarios~\citep{openai2023gpt4,team2023gemini,anthropic2024claude3,guo2025deepseekr1}. Financial institutions use LLMs for market analysis, healthcare providers employ them for clinical decision support, and agricultural organizations leverage them for crop management advice~\citep{wu2023bloomberggpt,singhal2023large,kuska2024ai}. However, these models often underperform in domain-specific settings where the language patterns, terminology, and knowledge needs differ from pre-training data~\citep{wu2023bloomberggpt,singhal2023large,gu2021domain,bella2024tackling,hu2025ttl}.

The standard responses include supervised fine-tuning (SFT) and reinforcement learning from human feedback (RLHF), which are effective when high-quality supervision is available~\citep{wei2022flan,ouyang2022instructgpt}. In production, however, collecting and refreshing domain-accurate data is costly, and specialized knowledge evolves over time, making maintenance difficult. Retrieval-augmented generation (RAG)~\citep{lewis2020rag,mao-etal-2021-generation} and few-shot prompting~\citep{an2023skill} mitigate the need for finetuning, but both rely on curated supervision in different forms: RAG requires maintained corpora, while prompting depends on carefully chosen examples. These methods alleviate but do not remove the reliance on explicit resources, motivating approaches that adapt without external supervision.

\begin{wrapfigure}{r}{0.48\columnwidth}
    \centering
    \vspace{-2em}
    \includegraphics[width=\linewidth,height=5.5cm,keepaspectratio]{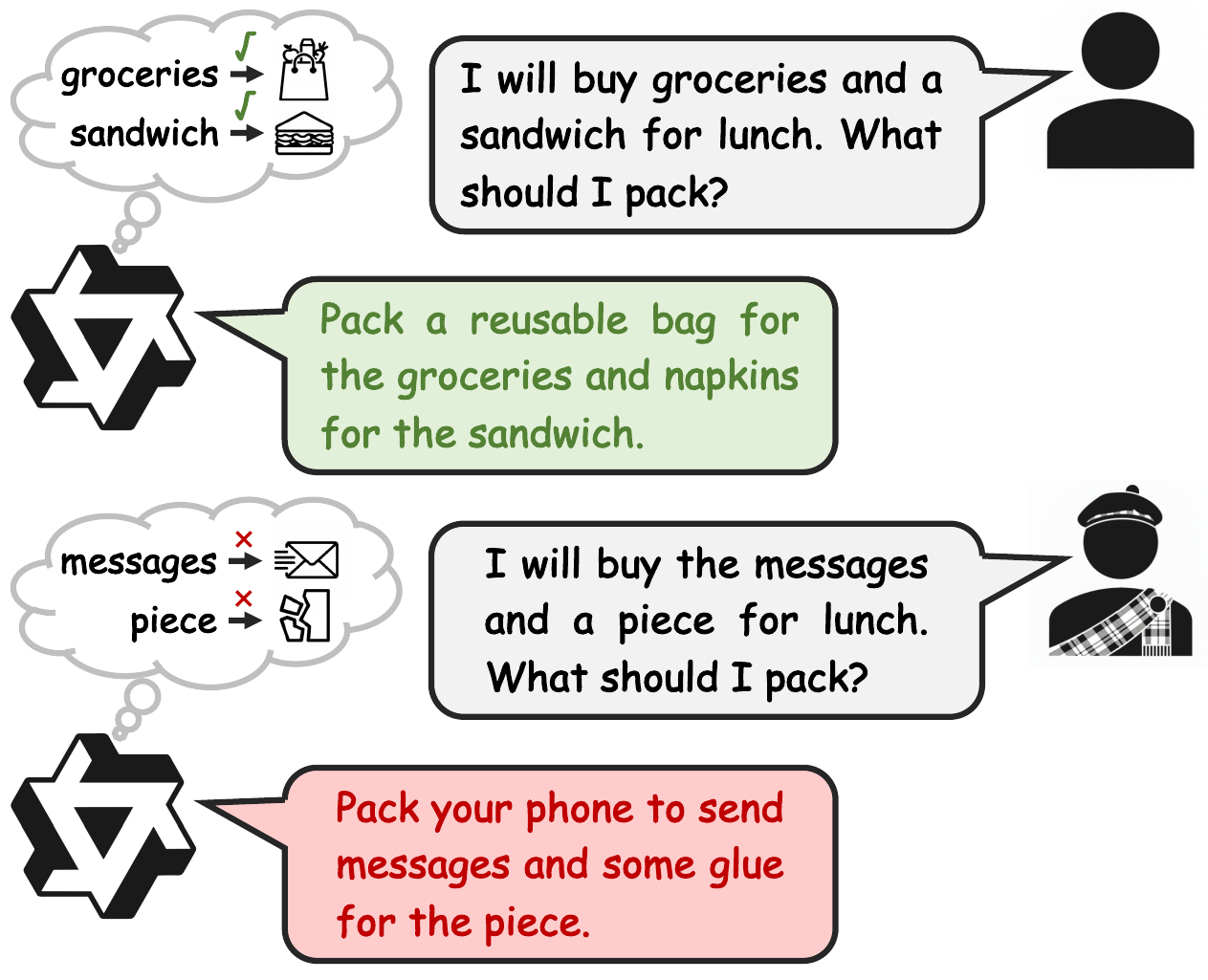}
    \vspace{-1.5em}
    \caption{\small{Illustration of LLM degradation under distribution shift: a Scottish dialect query (``messages and a piece'') is misinterpreted as unrelated intent.}}
    \label{fig:overview}
    \vspace{-1.25em}
\end{wrapfigure}

These constraints motivate a complementary direction: adapting models at inference time \emph{without external supervision}. Humans learn a language once and later adapt to new accents or dialects after brief exposure, without new explicit instruction, because the core vocabulary and grammar are already in place~\citep{clarke2004rapid,norris2003perceptual}. Analogously, LLMs possess broad base abilities from pre-training; they can still miss the intent of a question or fail to select the right knowledge, not because the knowledge is absent, but because query and answer distributions diverge from pre-training. For instance, as shown in Figure~\ref{fig:overview}, a query in Scottish dialect (“messages and a piece”) is misinterpreted by the model, even though the intended meaning is “groceries and a sandwich.” A human who already speaks English, however, can usually adapt after brief exposure to such dialectal variations and will eventually understand the phrase correctly. This mirrors the goal of test-time adaptation: adjusting to distribution shifts during inference without requiring new labeled supervision. For autoregressive LLMs, distribution shift yields measurable uncertainty patterns: domain-specific inputs trigger higher token-level perplexity, and decoding exhibits higher predictive entropy. Treating these quantities as self-supervised signals enables per-cohert adaptation under practical latency budgets. This converts deployment-time uncertainty into a training signal that narrows the train--deploy gap without labels.

Prior test-time adaptation for LLMs has typically optimized a single signal. Input-side objectives reduce perplexity to better match domain patterns~\citep{hu2025ttl}, yet they do not directly control decoding behavior. Output-side entropy minimization sharpens predictions~\citep{wang2021tent,niu2022eata}, but naive application to autoregressive generation can cause repetition and collapse~\citep{holtzman2020degeneration}. The challenge is to couple these signals so that the model becomes more confident and more domain-aware, while avoiding degeneration and unnecessary computation.

To this end, we propose Synergistic Test-time Adaptation (\method), a unified framework that couples input perplexity and output predictive entropy for LLMs. 
\method jointly reduces these uncertainties with guardrails that prevent degenerate text, and automatically allocates optimization effort to the dominant source of uncertainty per instance. The procedure is efficient: \method adapts with only 4--16 extra tokens per query and supports two deployment modes. The \modewocache mode updates during generation for maximum effect, while the \modecache mode pre-computes signals before decoding to reduce latency. Both modes are practical for real deployments.

Our contributions are as follows:
{\setlength{\leftmargini}{2.0em}
\begin{enumerate}
    \item We address the challenge of adapting LLMs to specialized domains under distribution shift and introduce Synergistic Test-time Adaptation (\method), a framework that jointly leverages input perplexity and output entropy as self-supervised signals to adapt LLMs without labeled data.
    \item We demonstrate consistent performance gains across domains and tasks on models spanning multiple families and parameter scales, while requiring only a small per-query token budget.
    \item We conduct extensive empirical analysis examining the effectiveness of different components across various scenarios, providing insights into when and how test-time adaptation benefits different types of distribution shifts.
\end{enumerate}}

%% file: tex/02_related_work.tex
\vspace{-0.75em}
\section{Related Works}
\vspace{-0.5em}

\noindent \textbf{Fine-tuning and retrieval from external knowledge.}
Supervised fine-tuning and instruction tuning improve performance for downstream tasks when high-quality labels or preferences are available~\citep{wei2022flan}, and RLHF aligns models with human feedback~\citep{ouyang2022instructgpt}. Retrieval-augmented methods combine parametric models with external corpora~\citep{lewis2020rag,guu2020realm}, but they introduce extra modules and costs. These approaches assume labeled data (SFT/RLHF) or a curated, queryable corpus (RAG), and are thus not directly applicable in our test-time setup, where only questions are given without labels or domain knowledge.

\noindent \textbf{Label-free test-time adaptation.}
Test-time adaptation updates models during inference without labels, aiming to mitigate performance degradation under distribution shift. In vision, entropy minimization (Tent) adapts classifier heads on unlabeled batches~\citep{wang2021tent}, with follow-ups improving stability and efficiency via sample selection~\citep{niu2022eata}, online adaptation~\citep{bar2024oem}, or conservative objectives~\citep{zhang2025cme}. For LLMs, test-time training with in-context examples improves few-shot reasoning~\citep{akyurek2024ttt}. Input-side updates with perplexity objectives also yield strong gains without labels~\citep{hu2025ttl}. These results highlight the utility of both input updates and output uncertainty control under shift.

\noindent \textbf{Reinforcement learning with verifiable or consistency signals.}
RLVR uses programmatic checks as reliable rewards~\citep{wen2025rlvr}. GRPO replaces the critic with group-based scoring~\citep{shao2024deepseekmath}, while variants like DAPO~\citep{yu2025dapo}, GFPO~\citep{shrivastava2025gfpo}, GSPO~\citep{zheng2025gspo}, and GVPO~\citep{zhang2025gvpo} address stability, efficiency, or length control.
Others explore test-time RL from consistency signals such as majority voting~\citep{zuo2025ttrl}, or simple entropy-based signals for math, code, and science tasks~\citep{agarwal2025ueem}. These methods rely on self-consistency or external verifiers, which limits their use in domain-specific or instruction tasks without reliable checkers. Our method is inspired by their stable optimization goals, but works without verifiers at test time.

%% file: tex/03_problem_setup.tex
\vspace{-0.5em}
\section{Problem Setup}
\vspace{-0.25em}
\label{sec:setup}

\subsection{Application Scenarios}
\vspace{-0.25em}
\label{sec:scenarios}

We investigate test-time adaptation for question answering under the challenging ``question-only'' condition, where the model is exposed to a large set of unlabeled questions from a shifted target distribution. The inputs are processed in batches, denoted by $X=\{x_j\}_{j=1}^M$. To adapt, the language model may generate a short prefix for each input. Crucially, the token budget for this prefix must be minimal to ensure that the adaptation process does not introduce significant latency, which would diminish its practical utility in real-world applications.

\paragraph{Cohort-Level Adaptation.}
Our setting resembles a multi-tenant model-as-a-service deployment. Before answering a batch window of target-domain questions $X$, the model performs a single self-supervised adaptation pass on the corresponding unlabeled pool. After this pass, parameters are frozen, and answers are generated for that cohort. We evaluate on this same cohort, which is a transductive test-time adaptation protocol where the unlabeled evaluation inputs are exactly those used for adaptation, and no ground-truth answers are accessed. When switching to a different domain, the model resets to a base snapshot, preventing cross-cohort information leakage or unintended accumulation. This workflow keeps inference lightweight while maintaining reliability across cohorts.

\vspace{-0.25em}
\subsection{Notations}
\vspace{-0.25em}
\label{sec:notation}

Let $x = (x_1, \dots, x_m)$ denote an input question, which is a sequence of $m$ tokens from a vocabulary $\mathcal{V}$. The corresponding response is a token sequence $y = (y_1, \dots, y_n)$ of length $n$. We denote the base LLM as $p_\theta$, parameterized by weights $\theta$. The model calculates the probability of a response $y$ given an input $x$ through an autoregressive factorization:
\begin{equation}
p_\theta(y \mid x) = \prod_{t=1}^{n} p_\theta(y_t \mid y_{<t}, x).
\end{equation}

During test-time adaptation, the model parameters are updated from $\theta$ to $\theta'$ based on the current input. Inference is then performed using the adapted model, $p_{\theta'}(\cdot \mid \cdot)$. For the adaptation step itself, the model generates a short prefix, denoted $\tilde{y}_{1:k}$, of length $k$. The value of $k$ also represents the extra token budget allocated for adaptation.

%% file: tex/04_method.tex
\vspace{-0.5em}
\section{Method: \method}
\vspace{-0.25em}
\label{sec:method}

\begin{figure*}[t]
    \centering
    \includegraphics[width=0.98\textwidth]{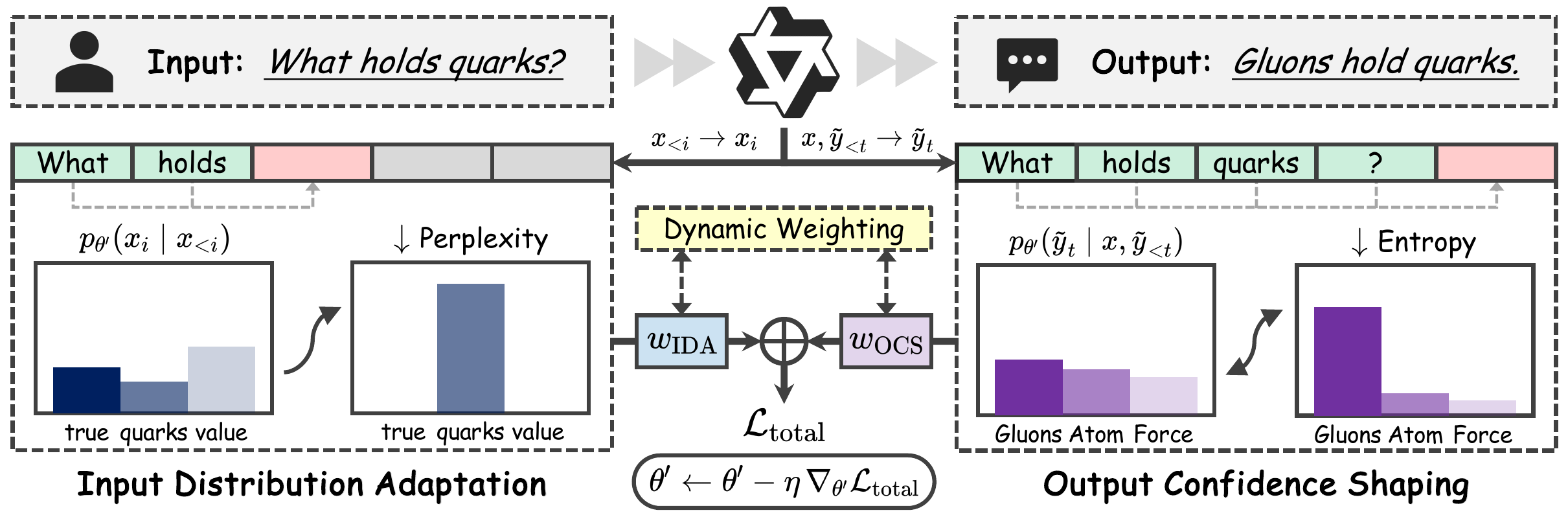}
    \caption{\small{Overview of the \method framework. \textit{Input Distribution Adaptation} lowers input perplexity, \textit{Output Confidence Shaping} reduces output entropy, and \textit{Dynamic Importance Weighting} balances the two signals. We leverage uncertainties as self-supervised signals for test-time adaptation.}}
    \label{fig:method_schematic}
    \vspace{-0.5em}
\end{figure*}

Our method, \method, realizes Synergistic Test-time Adaptation by coupling two complementary signals over a shared, short prefix context (Figure~\ref{fig:method_schematic}). \textit{Input Distribution Adaptation} pulls the input side toward the target domain by lowering the question’s perplexity; \textit{Output Confidence Shaping} pushes the output side toward confident yet anchored next-token distributions. These two signals act on the same prefix, and we coordinate them with a \textit{Dynamic Importance Weighting} rule that keeps their magnitudes comparable across instances. We elaborate on each of these components in the following sections. Additionally, we state the use of LLMs in Appendix~\ref{app:use_llm}.

\subsection{Input Distribution Adaptation}
\label{sec:ida}

To anchor the model in the target domain's specific language and concepts, we first optimize its ability to understand the incoming question $x$. Following recent test-time learning work~\citep{hu2025ttl}, \textit{Input Distribution Adaptation} minimizes prompt perplexity (equivalently NLL):
\begin{equation}
    \mathcal{L}_{\text{IDA}}(\theta') = -\frac{1}{m}\sum_{i=1}^{m} \log p_{\theta'}(x_i \mid x_{<i}).
\end{equation}
To focus adaptation on challenging instances, we employ a gating mechanism where the optimization is applied only to samples whose initial NLL under the base model $p_\theta$ exceeds a predefined threshold. For these selected samples, the loss is further amplified by a factor proportional to their NLL, promoting faster and more stable learning on difficult inputs.

\subsection{Output Confidence Shaping}
\label{sec:ocs}

While \textit{Input Distribution Adaptation} reduces input perplexity, it does not ensure coherent or confident generation. Models may still exhibit high predictive entropy or drift during decoding. To complement input-side adaptation, we introduce an output-oriented objective that regularizes the next-token distribution. Unlike many test-time procedures that update the model at every step, which risk error propagation and added cost, \method can decouple supervision from adaptation and explicitly shape outputs using entropy and reverse Kullback--Leibler (KL) terms.

For each input $x$, we form a short prefix of length $k$ and a reference distribution from the base model. Let
\begin{align}
\tilde{y}(x) &=
\begin{cases}
\textsc{GenPrefix}\!\left(p_\theta, x, k\right), & \text{\modecache mode},\\[2pt]
\textsc{GenPrefix}\!\left(p_{\theta'}, x, k\right), & \text{\modewocache mode},
\end{cases}
\end{align}

Given the generated prefix $\tilde{y}(x)$, we define the base-model reference logits at each step as
\begin{equation}
z_t^{\mathrm{ref}}(x) = \log p_\theta\!\left(\cdot \mid x,\tilde{y}_{<t}(x)\right),\quad t=1,\dots,k .
\end{equation}

In the \modecache mode, $\tilde{y}(x)$ and $\{z_t^{\mathrm{ref}}(x)\}_{t=1}^k$ are computed once with the frozen base model $p_\theta$ and cached for the whole adaptation. In the \modewocache mode, the model updates while generating its own short prefix; the base-model reference logits $\{z_t^{\mathrm{ref}}(x)\}$ are computed on the fly for the same context and are not cached.

The adapted model $p_{\theta'}$ conditions on $(x,\tilde{y}(x))$ with a base-model-forced forward pass to obtain learning signals. \textit{Output Confidence Shaping} then minimizes token-level predictive entropy along the prefix and regularizes the adapted distribution toward the base model. The entropy term aggregates next-token entropies,
\begin{equation}
    \mathcal{L}_{\text{ENT}}(\theta') \;=\; \sum_{t=1}^{k} H\!\left(p_{\theta'}(\cdot \mid x,\tilde{y}_{<t})\right),
\end{equation}
where $H(\cdot)$ is the Shannon entropy. We do not commit to a particular instantiation of the entropy computation here, leaving flexibility for implementation choices.

Test-time adaptation is known to be highly sensitive and can easily suffer from over-updating, which leads to model collapse. To prevent drift and collapse, we add a per-token reverse KL term~\cite{gu2023minillm} against the base-model reference,
\begin{equation}
    \mathcal{L}_{\text{KL}}(\theta') \;=\; \sum_{t=1}^{k} D_{\mathrm{KL}}\!\left(
        p_{\theta'}(\cdot \mid x,\tilde{y}_{<t})
        \,\Vert\,
        \mathrm{softmax}\!\big(z_t^{\mathrm{ref}}(x)\big)
    \right).
\end{equation}
The \textit{Output Confidence Shaping} objective combines these two parts,
\begin{equation}
    \mathcal{L}_{\text{OCS}}(\theta') \;=\; \mathcal{L}_{\text{ENT}}(\theta') \;+\; \lambda_{\text{KL}}\,\mathcal{L}_{\text{KL}}(\theta'),
\end{equation}
where $\lambda_{\text{KL}}$ balances confidence sharpening and proximity to the base model. The prefix length $k$ (typically $4$--$16$ tokens) sets the strength of the output-side signal relative to computation. Detailed discussions of the entropy objective design and the choice of the KL formulation are provided in Appendix~\ref{app:algorithm_design}.

\input{tex/main_algorithm}

\subsection{Dynamic Importance Weighting}
\label{sec:diw}

A static weighting between the \textit{Input Distribution Adaptation} objective and the \textit{Output Confidence Shaping} objective is suboptimal, because their relative difficulty varies across steps and instances. We therefore use a dynamic scheme that keeps the two contributions on a comparable scale, which helps stabilize training. The total loss is
\begin{equation}
    \mathcal{L}_{\text{total}}(\theta') \;=\; w_{\text{IDA}}^{(t)}\,\mathcal{L}_{\text{IDA}}(\theta') \;+\; w_{\text{OCS}}^{(t)}\,\mathcal{L}_{\text{OCS}}(\theta').
\end{equation}

\paragraph{Static baseline.}
As a point of reference, the static baseline fixes $w_{\text{IDA}}{=}w_{\text{OCS}}{=}1$.

\paragraph{Dynamic Loss-Ratio Weighting.}
To balance the two objectives, we propose a dynamic weighting scheme inspired by normalization-based methods in multi-task learning~\citep{chen2018gradnorm,liu2019end}. The core idea is to adjust each objective's weight at every step based on its current contribution to the total loss, while enforcing stability.

First, we track the overall loss magnitude using an exponential moving average (EMA) with momentum $\beta\in[0,1)$, which acts as a dynamic normalizer:
\begin{equation}
    \mathcal{L}^{(t)} = \beta\,\mathcal{L}^{(t-1)} + (1-\beta)\big(\mathcal{L}_{\text{IDA}}^{(t)} + \mathcal{L}_{\text{OCS}}^{(t)}\big).
\end{equation}
Using this normalizer, we compute the relative contribution of each loss, $r_i^{(t)} = \mathcal{L}_i^{(t)}/(\mathcal{L}^{(t)}+\varepsilon)$ for $i\in\{\text{IDA},\text{OCS}\}$, and normalize them to obtain preliminary weights $\tilde{w}_i^{(t)}=r_i^{(t)}/\sum_j r_j^{(t)}$. These are scaled by base coefficients $\lambda_i$, yielding $w_i^{(t)}=2\cdot \lambda_i\cdot \tilde{w}_i^{(t)}$.

However, we also observe that $L_\text{OCS}$ could always be orders of magnitude larger than $L_\text{IDA}$, where this ratio-based approach can cause training instability by effectively silencing one objective. To prevent this, we introduce a bounded rebalancing mechanism. We first clip the ratio of the two weights within a range 
\begin{equation}
\alpha^{(t)} \;\leftarrow\; 
\mathrm{clip}\!\left(
    \tfrac{w_{\text{OCS}}^{(t)}}{w_{\text{IDA}}^{(t)}},\,
    \alpha_{\min},\,\alpha_{\max}
\right).
\end{equation}

We then rescale the weights to maintain their sum, ensuring the total gradient magnitude remains controlled:
\begin{equation}
\big(w_{\text{IDA}}^{(t)},\,w_{\text{OCS}}^{(t)}\big) = \Big(\tfrac{2}{1+\alpha^{(t)}},\, \tfrac{2\,\alpha^{(t)}}{1+\alpha^{(t)}}\Big).
\end{equation}

This design keeps both objectives on a comparable scale. Clipping activates only under extreme loss imbalance to prevent dominance, and EMA-based normalization governs weighting. As the weights are set by forward-pass statistics rather than backpropagation, the EMA offers stability without requiring sensitive hyperparameter tuning (e.g., temperature). We compare our scheme with the static baseline in Section~\ref{sec:findings:diw}, and more details are shown in Appendix~\ref{app:hyperparameter_diw}.

\begin{wraptable}{r}{0.48\columnwidth}
\centering
\vspace{-2.5em}
\caption{\small{Adaptation cost during training.}}
\label{tab:complexity_wrap}
\vspace{0.5em}
\begin{tabular}{l c}
\toprule
\textbf{Method} & \textbf{Forward Passes} \\
\midrule
TENT, EATA & $(k+1)|\mathcal{D}|$ \\
TLM & $2|\mathcal{D}|$ \\
\method (\modewocache) & $(k+1)|\mathcal{D}|$ \\
\textbf{\method (\modecache)} & $\bm{|\mathcal{D}|}$ \\
\bottomrule
\end{tabular}
\vspace{-0.5em}
\end{wraptable}

\paragraph{Algorithm and Complexity.}
We summarize the computational cost of adaptation only during training in Table~\ref{tab:complexity_wrap}, comparing our approach with several baselines. The ``\method (\modecache)'' variant is notably efficient, requiring only a single forward pass per sample in the dataset ($\bm{|\mathcal{D}|}$), significantly outperforming current methods like TLM, TENT, and EATA. We refer to our method with prefix length $k$ as \method-$k$. The full procedure is in Algorithm~\ref{alg:method}. 
For each batch, adaptation runs a single base-model-forced forward pass of $p_{\theta'}$ over the length-$k$ prefixes in the \modecache mode, using cached base-model generation results and log-probabilities for the KL term. This removes repeated decoding and avoids feedback from unstable updates. Additionally, \modecache better exploits vLLM~\citep{kwon2023vllm} features, such as PagedAttention’s paged KV memory and continuous batching with prefix reuse, yielding faster training.

%% file: tex/main_algorithm.tex
\begin{algorithm}[t]
\caption{Training procedure of \method-$k$ with optional prefix cache}
\label{alg:method}
\small
\begin{algorithmic}[1]
\Require dataset $\mathcal{D}$, base model $p_{\theta}$, step size $\eta$, prefix len $k$, KL weight $\lambda_{\mathrm{KL}}$, mode (\textit{\modecache/\modewocache})
\Ensure adapted parameters $\theta'$
\If{mode = \modecache}
  \State Cache $\mathcal{C}=\{x \mapsto (\tilde y(x), z^{\mathrm{ref}}_{1:k}(x))\}$
  \algorithmiccomment{store prefix and reference logits}
\EndIf
\State $\theta' \leftarrow \theta$
\For{$s = 1$ \textbf{to} $S$}
  \State Sample mini-batch $\mathbf{X} \subset \mathcal{D},\ |\mathbf{X}|=B$
  \State Build tensors $\tilde{\mathbf{y}} \in \mathcal{Y}^B$, $Z^{\mathrm{ref}}_{1:k} \in \mathbb{R}^{B \times k \times V}$:
  \Statex \hspace{\algorithmicindent}\begin{minipage}[t]{0.45\linewidth}
    \fcolorbox{black}{gray!10}{%
    \begin{minipage}[t]{0.95\linewidth}
      \textbf{\modecache}\vspace{-0.75em}
      \begin{align*}
          \tilde{\mathbf{y}} &= (\tilde y(x))_{x \in \mathbf{X}},\\
          Z^{\mathrm{ref}}_{1:k} &= (z^{\mathrm{ref}}_{1:k}(x))_{x \in \mathbf{X}}
      \end{align*}
    \end{minipage}}
  \end{minipage}\hfill
  \begin{minipage}[t]{0.45\linewidth}
    \fcolorbox{black}{gray!10}{%
    \begin{minipage}[t]{0.95\linewidth}
      \textbf{\modewocache}\vspace{-0.75em}
      \begin{align*}
          \tilde{\mathbf{y}} &= \textsc{GenPrefix}(p_{\theta}, \mathbf{X}, k),\\
          Z^{\mathrm{ref}}_{1:k} &= \textsc{Logits}(p_{\theta}, \mathbf{X}, \tilde{\mathbf{y}}, k)
      \end{align*}
    \end{minipage}}
  \end{minipage}
  \State Run $p_{\theta'}(\mathbf{X}, \tilde{\mathbf{y}})$ 
        \algorithmiccomment{base-model-forced (\modecache) / generated prefix (\modewocache)}
  \State Compute losses: 
        $\mathcal{L}_{\mathrm{IDA}}, \mathcal{L}_{\mathrm{OCS}}, \mathcal{L}_{\mathrm{KL}} \in \mathbb{R}^B$
        \algorithmiccomment{Sec.~\ref{sec:ida}, Sec.~\ref{sec:ocs}}
  \State Compute weights: $w_{\mathrm{IDA}}, w_{\mathrm{OCS}} \in \mathbb{R}^B$ 
        \algorithmiccomment{Sec.~\ref{sec:diw}}
  \State Aggregate batch loss:
  $
  \mathcal{L}_{\mathrm{batch}}
  = \tfrac{1}{B}\Big( 
    \langle w_{\mathrm{IDA}}, \mathcal{L}_{\mathrm{IDA}} \rangle
    + \langle w_{\mathrm{OCS}}, \mathcal{L}_{\mathrm{OCS}} \rangle
    + \lambda_{\mathrm{KL}} \cdot \mathbf{1}^\top \mathcal{L}_{\mathrm{KL}} \Big)
  $
  \State Update: $\theta' \leftarrow \theta' - \eta \, \nabla_{\theta'} \mathcal{L}_{\mathrm{batch}}$
\EndFor
\Statex
\Return $\theta'$
\end{algorithmic}
\end{algorithm}

%% file: tex/05_experiments.tex
\vspace{-0.5em}
\section{Experiments}
\vspace{-0.25em}
\label{sec:experiments}

\vspace{-0.25em}
\subsection{Experimental Setup}
\vspace{-0.25em}
\label{sec:experiment_setup}

\paragraph{Datasets.} Following the experimental setup of~\citet{hu2025ttl}, we evaluate our method primarily on the \texttt{AdaptEval} benchmark suite, designed to test two key capabilities: downstream domain adaptation and instruction following. To this end, we use its two main components: \texttt{DomainBench} and \texttt{InstructBench}. \texttt{DomainBench} assesses model performance on specialized knowledge domains and comprises four datasets: \texttt{Agriculture}~\citep{kisanvaani2023agriculture}, \texttt{GeoSignal}~\citep{daven3_2023_geosignal}, \texttt{GenMedGPT}~\citep{wang2023genmedgpt}, and \texttt{Wealth}~\citep{bharti2023wealth}, while \texttt{InstructBench} measures the ability to adhere to diverse instructions and consists of three datasets: \texttt{Dolly}~\citep{Conover2023DollyV2}, \texttt{Alpaca-GPT4}~\citep{peng2023instruction}, and \texttt{InstructionWild}~\citep{ni2023instructionwild}. Additional details regarding each dataset are available in Appendix~\ref{app:datasets}.

\paragraph{Base Models and Baselines.} To validate the effectiveness and generalizability of our method, we conduct experiments using a diverse set of state-of-the-art open-source language models and compare against strong baselines. Our base models include the instruct version of \textsc{Llama 3.1-8B}~\citep{ai2024llama3_1}, \textsc{Llama 3.2-3B}~\citep{ai2024llama3_2}, and two instruct 
models from the Qwen series, \textsc{Qwen 2.5-7B} and \textsc{Qwen 2.5-14B}~\citep{qwen_team2024qwen2_5}. 

We compare \method{} against several methods: the base model without adaptation, which serves as a lower bound; TLM~\citep{hu2025ttl}, which adapts by optimizing input perplexity only; and two prominent methods from computer vision, Tent~\citep{wang2021tent} and EATA~\citep{niu2022eata}. Following the adaptations in~\citet{hu2025ttl}, we adapt their core principle of entropy minimization to the LLM's output distribution and implementation details to create strong baselines.

\paragraph{Evaluation Metrics.} We primarily use \rougelsum~\citep{lin2004rouge} to evaluate the quality of generated responses against the reference answers, capturing sentence-level overlap with summaries. A discussion of alternative metrics is provided in Appendix~\ref{app:other_metrics}.

\paragraph{Implementation Details.} We fine-tune models using Low-Rank Adaptation (LoRA)~\citep{hu2021lora} with a rank of 8, targeting the query and value projection matrices (\texttt{q\_proj} and \texttt{v\_proj}). Our implementation is based on the LLaMA Factory framework~\citep{zheng2024llamafactory}, with inference accelerated by the vLLM engine~\citep{kwon2023vllm}. For reproducibility, all responses are generated via greedy decoding. The training uses a learning rate of $1\times10^{-5}$, one epoch, and a cosine learning rate scheduler. Additional hyperparameters and details are provided in Appendix~\ref{app:implementation}.

\input{tex/main_table}

\vspace{-0.5em}
\subsection{Results}
\vspace{-0.25em}
\label{sec:results}

The main results are summarized in Table~\ref{tab:main_results}, showing that across all models and datasets, \method{} achieves clear improvements over both the base model and prior test-time adaptation methods. Entropy-only approaches such as Tent and EATA fail to adapt autoregressive LLMs, often collapsing performance to near-zero scores. Input-only perplexity optimization (TLM) is a stronger baseline and can be competitive in some cases, but it shows instability, including collapse on \texttt{Dolly}, and rarely delivers the best overall results. In contrast, \method{} combines input adaptation and output confidence shaping under dynamic weighting, yielding consistent and often state-of-the-art improvements across both \texttt{DomainBench} and \texttt{InstructBench}. The gains are particularly striking on some datasets; for example, on the \texttt{Agriculture} dataset, \method{} improves ROUGE-Lsum by over 120\% on \textsc{Qwen 2.5-7B} with only 4 extra tokens per query. The only notable exception is \texttt{GenMedGPT}, where TLM sometimes outperforms; we attribute this to its synthetic GPT-generated nature, whose distribution diverges from real clinical text and diminishes the value of shaping output confidence. Overall, \method{} significantly improves average \rougelsum, with gains of 40–60\% on \texttt{DomainBench} and 15–22\% on \texttt{InstructBench} depending on the model and variant. We also provide additional results under more prefix generation lengths in Table~\ref{tab:appendix_table_rougelsum} in the Appendix~\ref{app:experiments}.

\paragraph{Trends across model families and sizes.}
We observe that the relative gains vary systematically with the base model family. For the \textsc{Llama} family, larger models benefit more from \method{}, with the 8B variant showing larger relative improvements than the 3B variant. For the \textsc{Qwen} family, the opposite trend appears: the 7B model improves more than the 14B model. We hypothesize that this difference arises because \textsc{Qwen} models are more strongly post-trained and instruction-aligned, leaving less headroom for test-time adaptation.

%% file: tex/main_table.tex
\begin{table*}[t]
\small
\centering
\caption{\small{Main results on \texttt{DomainBench} and \texttt{InstructBench}. ROUGE-Lsum scores ($\times 100$ for visibility; higher is better). For each model and dataset, the highest score is \textbf{bold} and the second-highest is \underline{underlined}.}}
\vspace{0.5em}
\label{tab:main_results}
\resizebox{\textwidth}{!}{%
\begin{tabular}{l l *{4}{S} S *{3}{S} S}
\toprule
& & \multicolumn{5}{c}{\textbf{\texttt{DomainBench}}} & \multicolumn{4}{c}{\textbf{\texttt{InstructBench}}} \\
\cmidrule(lr){3-7} \cmidrule(lr){8-11}
\textbf{Model} & \textbf{Method} &
\texttt{Agriculture} & \texttt{GeoSignal} & \texttt{GenMedGPT} & \texttt{Wealth} &
\multicolumn{1}{c}{\textbf{Avg.}} &
\texttt{Dolly} & \texttt{Alpaca-GPT4} & \texttt{InstructWild} &
\multicolumn{1}{c}{\textbf{Avg.}} \\
\midrule

\multirow{10}{*}{\textsc{Llama-3.2-3B}}
& Base Model & 8.34 & 22.02 & 14.13 & 21.45 & 16.48 & 30.68 & 34.41 & 25.61 & 30.23 \\
& Tent~\citep{wang2021tent} & 0.98 & 4.59 & 9.32 & 2.33 & 4.30 & 5.66 & 5.72 & 6.41 & 5.93 \\
& EATA~\citep{niu2022eata} & 0.39 & 4.89 & 5.49 & 0.03 & 2.70 & 1.05 & 6.83 & 3.55 & 3.81 \\
& TLM~\citep{hu2025ttl} & 14.23 & 27.56 & \bestscore{24.29} & 26.73 & 23.20 & 24.77 & 37.66 & 27.66 & 30.03 \\
& \multicolumn{10}{>{\columncolor{gray!10}}c}{\modewocache} \\
& \method{}-4 & \secondscore{19.72} & 26.74 & 17.64 & \bestscore{29.10} & \secondscore{23.30} & 32.56 & \bestscore{40.53} & \secondscore{34.69} & \secondscore{35.93} \\
& \method{}-16 & 18.37 & 27.15 & 17.85 & 28.18 & 22.89 & 30.56 & \secondscore{39.72} & 32.08 & 34.12 \\
& \multicolumn{10}{>{\columncolor{gray!10}}c}{\modecache} \\
& \method{}-4 & \bestscore{20.12} & \bestscore{29.45} & 17.59 & \secondscore{29.07} & \bestscore{24.06} & \bestscore{34.12} & \bestscore{40.53} & \bestscore{36.15} & \bestscore{36.93} \\
& \method{}-16 & 15.38 & \secondscore{28.31} & \secondscore{19.66} & 28.28 & 22.91 & \secondscore{33.46} & 39.67 & 32.65 & 35.26 \\

\midrule

\multirow{10}{*}{\textsc{Llama-3.1-8B}}
& Base Model & 8.59 & 22.28 & 13.53 & 21.65 & 16.51 & 32.90 & 34.40 & 25.67 & 30.99 \\
& Tent & 1.16 & 3.79 & 0.74 & 13.22 & 4.73 & 0.45 & 4.84 & 9.78 & 5.02 \\
& EATA & 1.52 & 6.43 & 1.86 & 14.60 & 6.10 & 1.75 & 5.89 & 2.53 & 3.39 \\
& TLM & 16.33 & 28.85 & \secondscore{25.71} & 28.95 & 24.96 & 32.36 & 38.41 & 28.88 & 33.22 \\
& \multicolumn{10}{>{\columncolor{gray!10}}c}{\modewocache} \\
& \method{}-4 & \bestscore{20.17} & \secondscore{29.47} & \bestscore{26.48} & \secondscore{29.58} & \bestscore{26.43} & 34.61 & \bestscore{41.27} & \bestscore{36.15} & \bestscore{37.34} \\
& \method{}-16 & \secondscore{19.56} & 26.52 & 25.03 & 29.55 & \secondscore{25.16} & 32.98 & 39.45 & 35.05 & \secondscore{35.83} \\
& \multicolumn{10}{>{\columncolor{gray!10}}c}{\modecache} \\
& \method{}-4 & 16.49 & \bestscore{29.52} & 24.82 & 29.50 & 25.08 & \bestscore{35.45} & \secondscore{40.85} & \secondscore{35.71} & \bestscore{37.34} \\
& \method{}-16 & 15.17 & 29.19 & 21.23 & \bestscore{29.86} & 23.86 & \secondscore{35.42} & 39.85 & 32.08 & 35.78 \\

\midrule

\multirow{10}{*}{\textsc{Qwen-2.5-7B}}
& Base Model & 9.43 & 22.03 & 12.51 & 23.88 & 16.96 & 27.05 & 38.17 & 27.77 & 31.00 \\
& Tent & \secondscore{19.64} & 22.15 & 5.31 & 28.59 & 18.92 & 21.47 & 24.38 & 26.93 & 24.26 \\
& EATA & 16.30 & 21.24 & 12.83 & 22.57 & 18.23 & 30.57 & 27.01 & 23.81 & 27.13 \\
& TLM & 11.23 & 26.21 & \secondscore{29.67} & 28.13 & 23.81 & 31.05 & 43.08 & 30.76 & 34.96 \\
& \multicolumn{10}{>{\columncolor{gray!10}}c}{\modewocache} \\
& \method{}-4 & 17.68 & \secondscore{29.42} & \bestscore{29.74} & 29.69 & 26.63 & 35.69 & \secondscore{43.33} & 32.95 & 37.32 \\
& \method{}-16 & \bestscore{21.14} & 28.81 & 26.83 & \bestscore{30.25} & \secondscore{26.76} & 35.93 & 43.13 & 33.67 & 37.58 \\
& \multicolumn{10}{>{\columncolor{gray!10}}c}{\modecache} \\
& \method{}-4 & 19.40 & 29.37 & 29.56 & 29.67 & \bestscore{27.00} & \bestscore{36.51} & \bestscore{43.40} & \bestscore{34.07} & \bestscore{37.99} \\
& \method{}-16 & 18.31 & \bestscore{29.47} & 25.92 & \secondscore{29.79} & 25.87 & \secondscore{36.27} & 43.04 & \secondscore{33.72} & \secondscore{37.68} \\

\midrule

\multirow{10}{*}{\textsc{Qwen-2.5-14B}}
& Base Model & 10.67 & 23.46 & 14.42 & 24.36 & 18.23 & 28.06 & 39.34 & 28.12 & 31.84 \\
& Tent & 4.92 & 27.89 & 14.87 & 28.19 & 18.97 & 29.66 & 28.12 & 11.29 & 23.02 \\
& EATA & 1.88 & 28.23 & 3.17 & 27.97 & 15.31 & 22.99 & 26.08 & 25.33 & 24.80 \\
& TLM & 11.09 & 28.70 & \bestscore{32.20} & 29.48 & 25.37 & 34.04 & 42.20 & 30.59 & 35.61 \\
& \multicolumn{10}{>{\columncolor{gray!10}}c}{\modewocache} \\
& \method{}-4 & \secondscore{20.09} & \secondscore{30.57} & \secondscore{31.05} & \bestscore{30.14} & \bestscore{27.96} & \bestscore{37.04} & \bestscore{43.24} & 34.45 & \bestscore{38.24} \\
& \method{}-16 & 18.82 & 28.72 & 29.79 & \secondscore{29.95} & 26.82 & 35.57 & 42.86 & \bestscore{35.49} & 37.97 \\
& \multicolumn{10}{>{\columncolor{gray!10}}c}{\modecache} \\
& \method{}-4 & 19.52 & 30.45 & 28.91 & 29.53 & \secondscore{27.10} & \secondscore{36.32} & \secondscore{43.13} & 34.12 & 37.86 \\
& \method{}-16 & \bestscore{21.85} & \bestscore{30.93} & 22.26 & 29.57 & 26.15 & \bestscore{37.04} & 42.90 & \secondscore{34.46} & \secondscore{38.13} \\

\bottomrule
\end{tabular}
}
\end{table*}

%% file: tex/06_findings.tex
\vspace{-0.75em}
\section{Findings based on \method}
\vspace{-0.5em}
\label{sec:findings}
\begin{figure*}[t]
    \centering
    \includegraphics[width=0.98\textwidth]{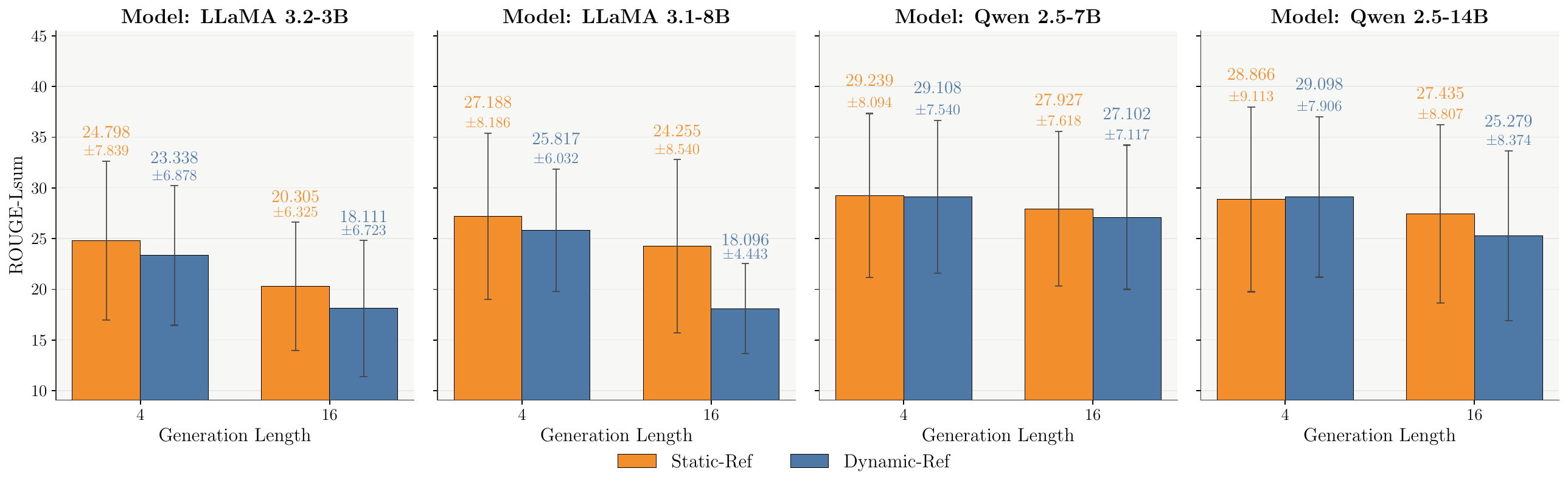}
    \vspace{-0.5em}
    \caption{\small{\rougelsum scores under different generation lengths (4 vs. 16) and models. Results are shown for both \modecache \textcolor[RGB]{242,142,43}{\rule{7pt}{7pt}} and \modewocache \textcolor[RGB]{78,121,167}{\rule{7pt}{7pt}}, with error bars indicating standard deviations.}}
    \label{fig:finding1_2}
    \vspace{-1.25em}
\end{figure*}

In this section, we analyze the design choices of \method{} by addressing research questions that are central to understanding its performance and robustness. Specifically, we investigate:
\vspace{-0.5em}
{\setlength{\leftmargini}{2.0em}
\begin{itemize}
    \item \textbf{Q1:} Is longer prefix generation always better for adaptation?
    \item \textbf{Q2:} Can the computational efficiency of the \modecache mode be maintained without sacrificing performance?
    \item \textbf{Q3:} Does Kullback–Leibler (KL) divergence genuinely contribute to model stability?
    \item \textbf{Q4:} Is Dynamic Importance Weighting necessary for balancing input and output objectives?
\end{itemize}}

\begin{wrapfigure}{r}{0.48\columnwidth}
    \centering
    \vspace{-3.25em}
    \includegraphics[width=\linewidth,height=5.5cm,keepaspectratio]{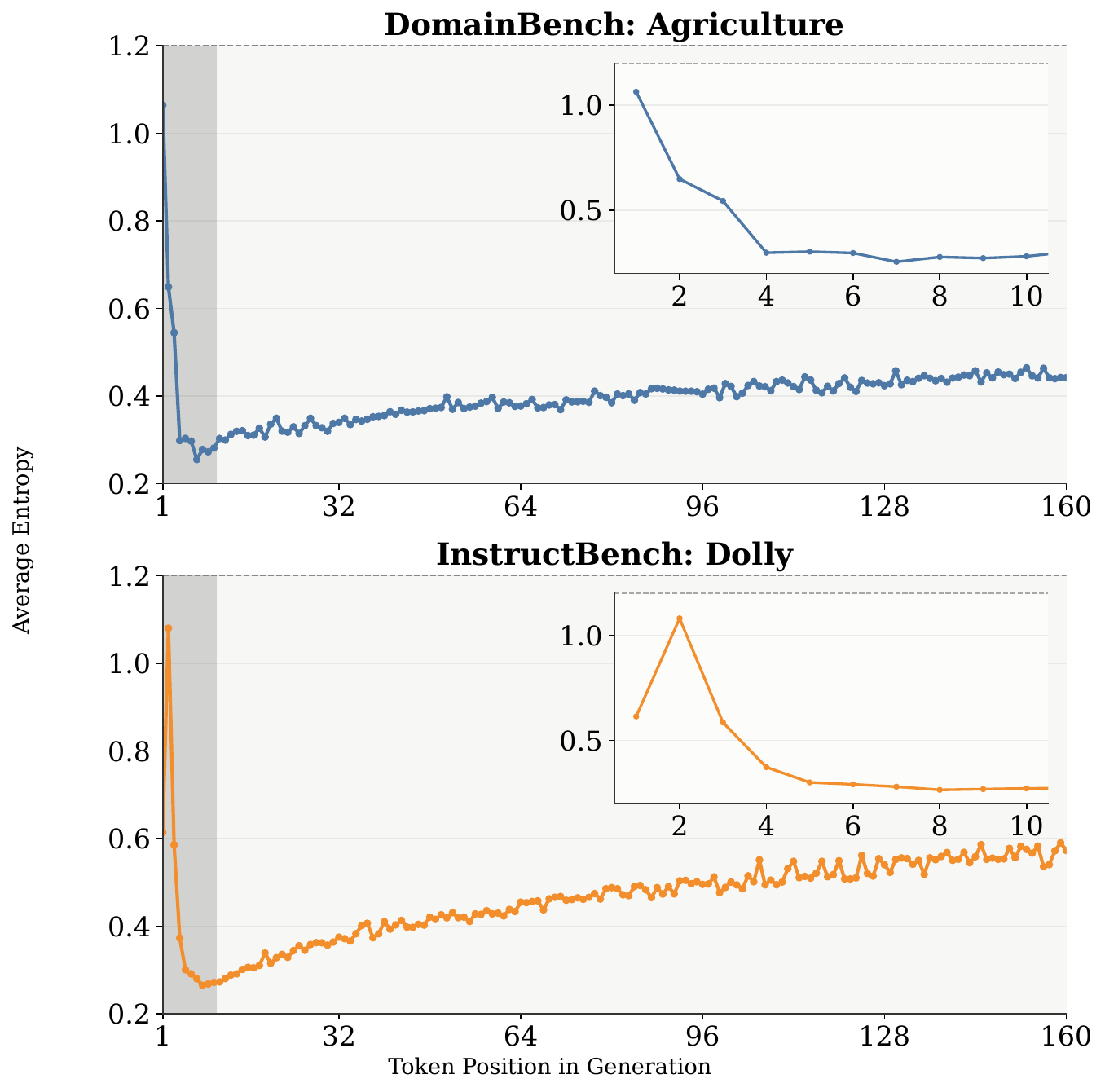}
    \vspace{-1.0em}
    \caption{\small{Average token-level response entropy computed by averaging across all responses.}}
    \label{fig:average_token_entropy}
    \vspace{-0.75em}
\end{wrapfigure}

\vspace{-0.5em}
\subsection{\texorpdfstring{Impact of Prefix Generation Length ($k$)}{Impact of Prefix Generation Length (k)}}
\vspace{-0.5em}
\label{sec:findings:prefix_length}

We first average results across tasks and then aggregate by model and generation length (marginalizing over update modes) to obtain Figure~\ref{fig:finding1_2}. Across all base models, a short prefix ($k=4$) outperforms a longer prefix ($k=16$). The average gains range from small but consistent (about +1 \rougelsum point on \textsc{Qwen 2.5-7B}) to more pronounced improvements (about +6 points on \textsc{LLaMA 3.2-3B}). This pattern indicates that most of the useful adaptation signal is contained in the earliest few tokens, while extending the prefix primarily increases variance and susceptibility to incidental noise without providing commensurate benefit. Consequently, $k=4$ offers a better stability–efficiency trade-off and is a robust default for adaptation. As shown in Fig.~\ref{fig:average_token_entropy}, the token-level response entropy on two representative datasets follows the same pattern: a very high spike at the first few tokens, followed by a rapid drop and a stable range for the rest of the generation. For both the domain-specific instruction-following set, the maximum occurs around $k\approx4$. This is consistent with our findings and supports adapting on the first few high-entropy tokens (e.g., $k=4$), which carry most of the useful signal. In contrast, longer prefixes mainly add noise and can lead to overfitting with little additional benefit.

\vspace{-0.5em}
\subsection{\modecache{} vs.\ \modewocache{} for Deployment}
\vspace{-0.5em}
\label{sec:findings:cache_modes}

To isolate the effect of update mode, we average results across generation lengths and compare \modecache{} with \modewocache{} in Figure~\ref{fig:finding1_2}. \modecache{} is consistently more stable across models and, on average, performs as well as or better than \modewocache{}. The advantage is clear in the \textsc{LLaMA} family (e.g., about +5 point \rougelsum on \textsc{LLaMA 3.1-8B} when averaged across lengths), while the gap is smaller in the \textsc{Qwen} family (e.g., less than +1 point on \textsc{Qwen 2.5-7B}). We attribute the reduced gap in \textsc{Qwen} to stronger post-training that makes online updates less sensitive to prefix drift. Considering stability and cost (one forward pass per sample for \modecache{}), \modecache{} is the recommended default for practical deployment, with \modewocache{} reserved for scenarios that explicitly benefit from tight coupling to the live decoding trajectory.

\begin{wrapfigure}{r}{0.48\columnwidth}
    \centering
    \vspace{-2em}
    \includegraphics[width=0.98\linewidth]{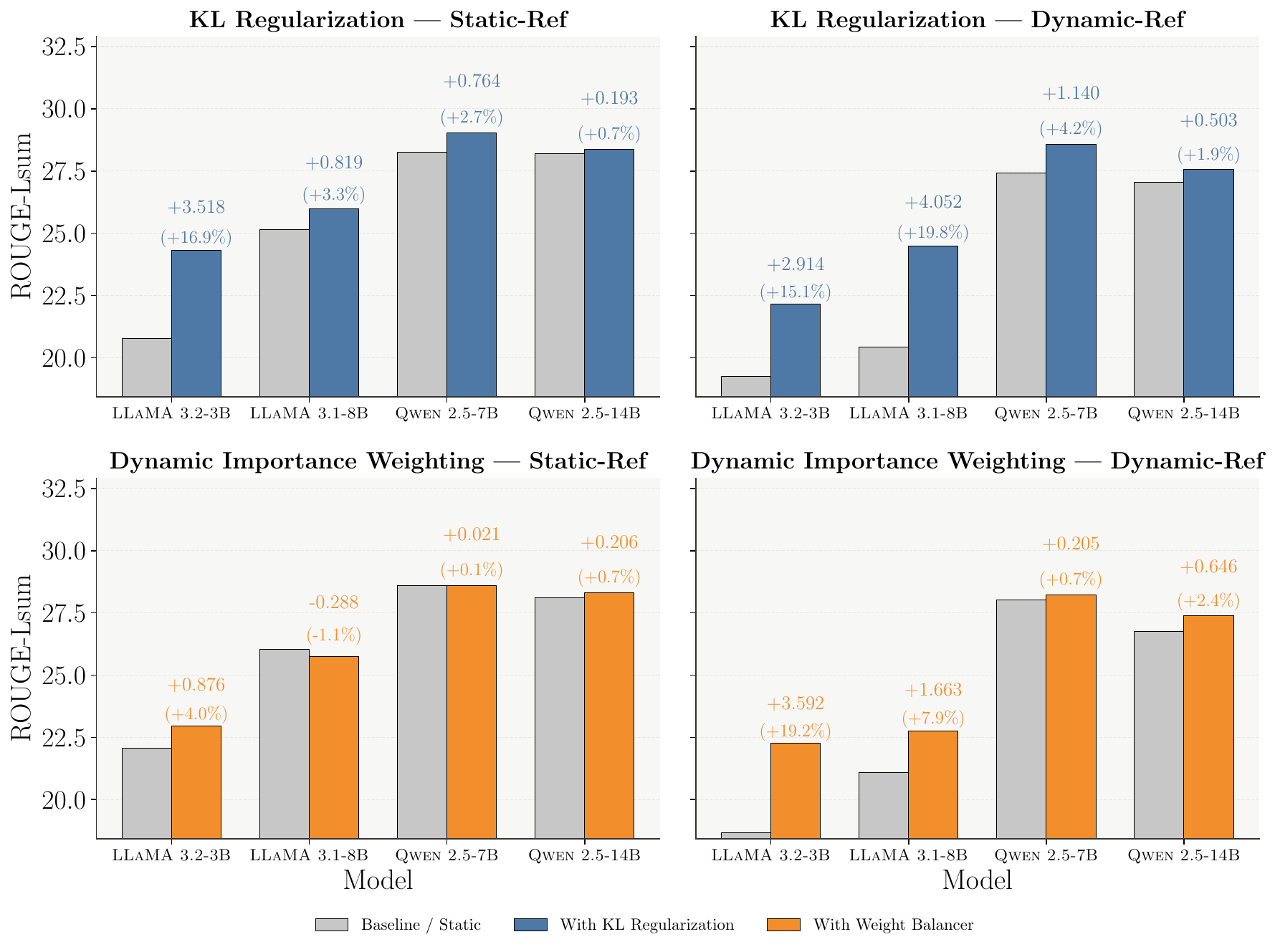}
    \vspace{-1em}
    \caption{\small{Ablations of KL regularization \textcolor[RGB]{78,121,167}{\rule{7pt}{7pt}} and \textit{Dynamic Importance Weighting} \textcolor[RGB]{242,142,43}{\rule{7pt}{7pt}} on \rougelsum across models. Both absolute and relative improvements (\%) are shown.}}
    \label{fig:finding3_4}
    \vspace{-2em}
\end{wrapfigure}

\vspace{-0.75em}
\subsection{Role of KL Divergence for Stability}
\vspace{-0.5em}
\label{sec:findings:kl_divergence}

We compare \method{} with and without a KL term that penalizes divergence from the base policy during online updates. To isolate this factor, we average across generation lengths and report results by model family and update mode. In Fig.~\ref{fig:finding3_4} (blue bars), enabling KL shows two consistent effects. First, it yields larger gains in \modewocache than in \modecache{}. A useful view is to treat the KL term as a trust-region: it restricts the adaptation to stay close to the base model, preventing abrupt shifts caused by transient gradients during decoding. This constraint is more important for \modewocache{}, where the model updates with the dynamic references and small errors can accumulate; \modecache{} uses a fixed reference, so drift is naturally smaller. Second, it more or less improves the average \rougelsum across models. The improvement is clearer in the \textsc{LLaMA} family, while the \textsc{Qwen} family shows smaller but steady gains, which we attribute to stronger post-training that already constrains the adaptation. Further details are in Appendix~\ref{app:kl}

\vspace{-0.75em}
\subsection{Necessity of Dynamic Importance Weighting}
\vspace{-0.5em}
\label{sec:findings:diw}

We ablate \textit{Dynamic Importance Weighting} by comparing it to a fixed weighting while holding other settings constant. We average across generation lengths and report by model family and update mode. In Fig.~\ref{fig:finding3_4} (orange bars), our scheme improves \rougelsum on most models. The net gain is larger under \modewocache{} than under \modecache{}, because the evolving reference amplifies sensitivity to early-token updates and DIW offsets this by rebalancing gradients on the fly. The effect is more pronounced for the \textsc{LLaMA} family, while \textsc{Qwen} shows smaller but consistent gains, which we attribute to stronger post-training that already reduces conflicts between objectives.

Mechanistically, DIW keeps the \textit{Input Distribution Adaptation} and \textit{Output Confidence Shaping} losses on a comparable scale using EMA-normalized loss ratios with a clipped weight ratio, which prevents one objective from dominating when their magnitudes differ by orders. In addition, it also alleviates the inherent instability caused by the non-smooth entropy patterns of response tokens (see Figure~\ref{fig:average_token_entropy}).
DIW and KL are complementary: KL limits adaptation drift, and DIW balances the two objectives step by step. For deployment, we enable DIW with KL by default. Adding DIW adds robustness to long generations and mixed-domain workloads without extra inference cost.

%% file: tex/07_conclusion.tex
\vspace{-1em}
\section{Conclusion}
\vspace{-0.5em}
\label{sec:conclusion}
This study introduces Synergistic Test-time Adaptation(\method{}), a novel label-free framework that adapts LLMs to specialized domains or scenarios at inference time. By synergistically coupling input perplexity and output entropy, \method{} provides a more robust and effective solution to distribution shifts than current approaches, improving both domain awareness and generation stability.
Experiments and findings demonstrate that \method{} can consistently improve performances across diverse models and benchmarks, delivering substantial gains with minimal computational overhead. This framework enhances the optimization of LLM deployment for both efficiency and reliability, promising a practical path for adaptation in label-scarce specialized domains. Future work will focus on extending this synergistic principle to more diverse generative tasks and deployment scenarios.

%% file: tex/99_appendix.tex
\appendix
\section{Appendix}
\etocsetnexttocdepth{3}    %
\localtableofcontents      %

\subsection{Details of Algorithm Design}
\label{app:algorithm_design}
This section clarifies our design choices with notation consistent with Section~\ref{sec:method}, covering the entropy objective (cumulative versus average) and why we use reverse KL divergence instead of forward KL.

\subsubsection{Entropy Objective}
\label{app:entropy_objective}
Recall that \textit{Output Confidence Shaping} aggregates token-level entropies along the length-$k$ prefix. We consider two variants that are compatible with the notation in Section~\ref{sec:ocs}. Let
\[
p_t'(\cdot) \;=\; p_{\theta'}\!\left(\cdot \mid x,\tilde{y}_{<t}\right),\quad t=1,\dots,k .
\]
The \emph{cumulative} form sums the entropies over the prefix,
\begin{equation}
\mathcal{L}_{\text{ENT}}^{\text{cum}}(\theta') \;=\; \sum_{t=1}^{k} H\!\big(p_t'(\cdot)\big) ,
\label{eq:cum-ent}
\end{equation}
while the \emph{average} form normalizes by $k$,
\begin{equation}
\mathcal{L}_{\text{ENT}}^{\text{avg}}(\theta') \;=\; \frac{1}{k}\sum_{t=1}^{k} H\!\big(p_t'(\cdot)\big) .
\label{eq:avg-ent}
\end{equation}
When losses are combined with fixed coefficients, the two forms differ only by a constant factor. In our setting, however, the absolute scale interacts with \textit{Dynamic Importance Weighting} (Section~\ref{sec:diw}), which uses forward-pass magnitudes to rebalance objectives. Empirically, the cumulative form in~\eqref{eq:cum-ent} gives a stronger and more stable signal, leading to small but consistent gains across models and datasets. We therefore use $\mathcal{L}_{\text{ENT}}^{\text{cum}}$ in all main results.

\subsubsection{KL Regularization Details}
\label{app:kl_details}
Let the base-model reference distribution at step $t$ be
\[
p_t^{\mathrm{ref}}(\cdot) \;=\; \mathrm{softmax}\!\big(z_t^{\mathrm{ref}}(x)\big) \;=\; p_{\theta}\!\left(\cdot \mid x,\tilde{y}_{<t}\right).
\]
We consider two KL choices between the adapted distribution $p_t'(\cdot)$ and the reference $p_t^{\mathrm{ref}}(\cdot)$.

\paragraph{Forward KL (mode-covering).}
\begin{equation}
\mathcal{L}_{\text{KL}}^{\text{fwd}}(\theta') \;=\; \sum_{t=1}^{k} D_{\mathrm{KL}}\!\left(
p_t^{\mathrm{ref}} \,\Vert\, p_t'
\right)
\;=\; \sum_{t=1}^{k}\sum_{a\in\mathcal{V}} p_t^{\mathrm{ref}}(a)\,\log\frac{p_t^{\mathrm{ref}}(a)}{p_t'(a)} .
\label{eq:fwd-kl}
\end{equation}
This penalizes the adapted model for \emph{missing} probability mass where the reference has support, encouraging coverage of all reference modes.

\paragraph{Reverse KL (mode-seeking).}
\begin{equation}
\mathcal{L}_{\text{KL}}^{\text{rev}}(\theta') \;=\; \sum_{t=1}^{k} D_{\mathrm{KL}}\!\left(
p_t' \,\Vert\, p_t^{\mathrm{ref}}
\right)
\;=\; \sum_{t=1}^{k}\sum_{a\in\mathcal{V}} p_t'(a)\,\log\frac{p_t'(a)}{p_t^{\mathrm{ref}}(a)} .
\label{eq:rev-kl}
\end{equation}
Reverse KL has a well-known mode-seeking behavior: to reduce~\eqref{eq:rev-kl}, the adapted distribution concentrates on high-density regions of $p_t^{\mathrm{ref}}$; if $p_t'(a)>0$ while $p_t^{\mathrm{ref}}(a)\approx 0$, the penalty becomes very large, discouraging exploration of regions that the reference assigns negligible probability to. This property is desirable in our test-time setting, where supervision is absent and unstable on-the-fly updates can drift. Using reverse KL, therefore, acts as a practical trust region that anchors the adapted model to the base policy while still allowing entropy reduction on the prefix.

\paragraph{Choice in \method.}
We adopt the reverse form in~\eqref{eq:rev-kl} and use it in the \textit{Output Confidence Shaping} objective
\[
\mathcal{L}_{\text{OCS}}(\theta') \;=\; \mathcal{L}_{\text{ENT}}^{\text{cum}}(\theta') \;+\; \lambda_{\text{KL}}\,\mathcal{L}_{\text{KL}}^{\text{rev}}(\theta') .
\]
Forward KL in~\eqref{eq:fwd-kl} is more tolerant of spreading mass and can encourage mode coverage, which raises entropy and reduces stability during online updates. In contrast, reverse KL provides stronger safeguards against degenerate repetition and off-support drift, especially in \modewocache{} where references evolve with the prefix. In all experiments, we use reverse KL; ablations in Appendix~\ref{sec:findings:kl_divergence} show that it improves robustness and average performance.

\subsection{Details of Datasets}
\label{app:datasets}
Here we describe the benchmarks and datasets used in our experiments, primarily derived from \texttt{AdaptEval}~\citep{hu2025ttl}. We adopt the original data splits and preprocessing protocols established in the benchmark.

\subsubsection{DomainBench}
\texttt{DomainBench} evaluates model adaptation in specialized domains requiring factual precision and domain-specific reasoning. It spans \texttt{Agriculture}, \texttt{Geography}, \texttt{GenMedGPT}, and \texttt{Wealth}, totaling over 110k examples. These datasets jointly test whether models can move beyond everyday text and produce reliable, domain-specific responses.

\paragraph{\texttt{Agriculture}.} The \texttt{Agriculture} dataset~\citep{kisanvaani2023agriculture} includes 22.6k Q\&A pairs on soil, crop growth, irrigation, fertilizer use, pest control, and weather effects. Questions are posed in a practical, farmer-oriented style, while answers provide concise, actionable guidance. It evaluates whether models can capture applied agricultural knowledge and generate context-appropriate recommendations.

\paragraph{\texttt{GeoSignal}.} The \texttt{GeoSignal} dataset~\citep{daven3_2023_geosignal} contains 39.7k instructions spanning mineral classification, stratigraphic analysis, tectonic features, and geospatial terms. It blends general tasks with domain-specific reasoning, such as relation inference and fact checking. The dataset challenges models to handle professional geoscientific language and structured knowledge.

\paragraph{\texttt{GenMedGPT}.} The \texttt{GenMedGPT} dataset~\citep{wang2023genmedgpt} is a synthetic medical corpus of 5.5k patient–doctor dialogues. Patient queries describe symptoms or conditions, and responses emulate clinical advice across diagnostics, pharmacology, and lifestyle guidance. It tests whether models can adapt to medical discourse and emulate expert consultation.

\paragraph{\texttt{Wealth}.} The \texttt{Wealth} dataset~\citep{bharti2023wealth} provides over 44k instructions on finance, covering accounting, taxation, market analysis, and investment strategies. Prompts follow an Alpaca-style format with short instructions and extended answers. It measures a model’s ability to reason about financial concepts and generate coherent domain-specific responses.

\subsubsection{InstructBench}
\texttt{InstructBench} assesses general instruction-following ability across curated and naturally occurring prompts. It combines datasets of different sizes and styles to test robustness, adaptability, and generalization in open-ended instruction adherence.

\paragraph{\texttt{Dolly}.} The \texttt{Dolly} dataset~\citep{Conover2023DollyV2} contains 15k human-authored instructions spanning brainstorming, classification, QA, summarization, and information extraction. All responses are concise and practical, reflecting workplace and educational use cases. It serves as a strong baseline for evaluating general instruction-following quality.

\paragraph{\texttt{Alpaca-GPT4}.} The \texttt{Alpaca-GPT4} dataset~\citep{peng2023instruction} consists of 52k instructions paired with GPT-4 responses. The dataset covers explanation, summarization, multi-step reasoning, and procedural tasks. Its detailed and fluent answers allow testing of whether models can follow complex instructions and maintain coherence across longer generations.

\paragraph{\texttt{InstructionWild}.} The \texttt{InstructionWild} dataset~\citep{ni2023instructionwild} includes over 110k real-world prompts collected from social media, open-source communities, and forums. The instructions are highly diverse, often noisy, and context-rich, ranging from casual conversational queries to technical tasks. It provides a challenging benchmark for robustness to non-curated, long-tail instructions.

\subsection{Details of Experiments}
\label{app:experiments}

We also provide in Table~\ref{tab:appendix_table_rougelsum} a more detailed version of the main results reported in Table~\ref{tab:main_results}. From these results, we observe that using a prefix generation length of $k=4$ yields the best overall performance.

\paragraph{\rougelsum as the main evaluation metric.} 
\rougelsum~\citep{lin2004rouge} measures the longest common subsequence (LCS) between a generated sequence and a reference summary, aggregating precision and recall over the subsequence. Unlike other ROUGE variants, \rougelsum operates at the sentence level, which makes it particularly suited for summarization-style evaluation, as it rewards long, in-order matches while allowing gaps. Higher \rougelsum{} indicates closer preservation of reference content and structure.

\subsubsection{Other Evaluation Metrics}
\label{app:other_metrics}

\paragraph{BERTScore-F1.}
BERTScore-F1~\cite{zhang2019bertscore} measures semantic similarity using contextual embeddings. We compute it with the official \textttsplit{bert-score}\footnote{\url{https://github.com/Tiiiger/bert_score}} implementation and the \textttsplit{bert-base-multilingual-cased}\footnote{\url{https://huggingface.co/google-bert/bert-base-multilingual-cased}} model. To control length effects, both prediction and reference are tokenized and truncated to at most 500 tokens, after which the two strings are trimmed to the same length before scoring. We report the mean F1 across all examples. The detailed results are shown in Table~\ref{tab:appendix_table_bertscoref1}.

\paragraph{ROUGE-1.}
ROUGE-1~\cite{lin2004rouge} measures unigram overlap. We use \textttsplit{rouge\_score.RougeScorer}\footnote{\url{https://pypi.org/project/rouge-score/}} with \textttsplit{use\_stemmer=True} and \textttsplit{split\_summaries=True}. As preprocessing, we replace ``\textttsplit{<n>}'' with a space, segment the candidate into sentences with \textttsplit{nltk}\footnote{\url{https://github.com/nltk/nltk}}, and join sentences with newlines. Scores are the F1 variant averaged over examples. The detailed results are shown in Table~\ref{tab:appendix_table_rouge1}.

\paragraph{ROUGE-2.}
ROUGE-2 extends the overlap to bigrams, capturing short phrase consistency. We use the same \textttsplit{rouge\_score} settings as for ROUGE-1 (\textttsplit{use\_stemmer=True}, \textttsplit{split\_summaries=True}) and the same sentence-level preprocessing (\textttsplit{<n>} replacement, sentence segmentation, newline joins). We report F1 averaged over examples. The detailed results are shown in Table~\ref{tab:appendix_table_rouge2}.

\paragraph{ROUGE-L.}
ROUGE-L computes the longest common subsequence overlap, rewarding in-order matches while allowing gaps. Implementation and preprocessing follow the same protocol as above (\textttsplit{rouge\_score} with \textttsplit{use\_stemmer=True}, \textttsplit{split\_summaries=True}; sentence segmentation and newline joins). We report the F1 variant averaged over examples. The detailed results are shown in Table~\ref{tab:appendix_table_rougel}.

\paragraph{BLEU.}
BLEU~\cite{papineni2002bleu} is based on modified $n$-gram precision with a brevity penalty. We compute sentence-level BLEU using the NLTK implementation with the standard smoothing method~4. Tokenization uses the NLTK word tokenizer, with a whitespace fallback if the tokenizer is unavailable. We average the sentence-level scores over examples. The detailed results are shown in Table~\ref{tab:appendix_table_bleu}.

\paragraph{Summary.}
Across the tables, we observe that \method consistently achieves strong performance across multiple metrics with different prefix generation length $k$. Given that \rougelsum is more robust, we highlight in the main results (Table~\ref{tab:main_results}) the configurations with $k=4$ and $k=16$, which stand out in Table~\ref{tab:appendix_table_rougelsum}.

\subsection{Details of Findings}
\label{app:finding}
\label{app:kl}
\subsubsection{KL Configuration}
In practice, a moderate KL coefficient works well. For the \textsc{Qwen} family with stronger post-training, larger models benefit from a smaller KL coefficient since their lower-entropy outputs make the same KL weight overly restrictive; for the \textsc{LLaMA} family, we keep a single coefficient across sizes. We enable KL by default and tune it following these family-specific rules. Concretely, we set the KL coefficient to $0.16$ for all \textsc{LLaMA} models, while for \textsc{Qwen-2.5}, the $7$B variant uses $0.16$ and the $14$B variant uses $0.01$.

\subsubsection{Hyperparameters of Dynamic Importance Weighting}
\label{app:hyperparameter_diw}

We adopt a dynamic importance weighting strategy with three hyperparameters: 
an EMA decay coefficient $\beta$ (default $0.9$) to smooth the total loss, 
a lower bound $\text{floor}$ (default $10^{-3}$), 
and an upper bound $\text{ceil}$ (default $10^{3}$) to constrain the loss ratio and prevent extreme imbalance. 
These values are fixed in all reported experiments.

\subsection{Details of Implementation}
\label{app:implementation}
Our experiments were conducted on high-performance servers equipped with either four or six NVIDIA A800 GPUs (80GB memory each) or eight NVIDIA H100 GPUs (80GB memory each). 
The A800 machines with four GPUs used the SXM4 version, while those with six GPUs were configured with the PCIe version. 
All systems were built with Intel(R) Xeon(R) Platinum CPUs, 1TB of RAM, and a software environment consisting of Python~3.11, PyTorch~2.4, and NCCL~2.21.5 to ensure reproducibility.

\subsection{The Use of Large Language Models}
\label{app:use_llm}
We acknowledge the use of a Large Language Model (LLM) to assist with language editing and polishing of this manuscript. The LLM's role was strictly limited to improving grammar, clarity, and phrasing. All scientific ideas, methodologies, results, and conclusions presented herein are the original work of the authors. The authors have thoroughly reviewed all revisions and assume complete responsibility for the entirety of the paper's content.

\subsection{Ethics Statement}
\label{app:ethics}
We affirm compliance with the ICLR Code of Ethics. Our work studies label-free test-time adaptation using publicly available benchmarks and does not involve new data collection, human subjects, or personally identifiable information. We follow the licenses of all datasets and base models cited in the paper. Because some tasks touch on finance, medicine, and agriculture, model outputs may carry risk if taken as advice. Our experiments are research-only; the method is not intended for clinical or financial decision-making without qualified human oversight. Deployments should include content filters, disclaimers, and domain-expert review, and must comply with local laws and institutional policies. We report no conflicts of interest or external sponsorship that could bias the results. The computational overhead is small (4--16 extra tokens per query and one forward pass per sample in \modecache), which limits environmental impact relative to standard fine-tuning.

\subsection{Reproducibility Statement}
\label{app:reproducibility}
We aim to make the work reproducible. The method is fully specified in Section~\ref{sec:method} with pseudocode in Algorithm~\ref{alg:method}. Datasets, splits, and preprocessing follow Appendix~\ref{app:datasets}. Training and inference settings, including LoRA configuration, learning schedules, gating, KL weighting, and decoding, are detailed in Appendix~\ref{app:implementation}; KL details are in Appendix~\ref{app:kl}.

\clearpage

\input{tex/appendix_table_rougelsum}

\input{tex/appendix_table_bertscore}

\input{tex/appendix_table_rouge1}

\input{tex/appendix_table_rouge2}

\input{tex/appendix_table_rougel}

\input{tex/appendix_table_bleu}

%% file: tex/appendix_table_rougelsum.tex
\begin{table*}[t]
\small
\centering
\caption{\small{Detailed results on \texttt{DomainBench} and \texttt{InstructBench}. ROUGE-Lsum scores ($\times 100$ for visibility; higher is better). For each model and dataset, the highest score is \textbf{bold} and the second-highest is \underline{underlined}.}}
\label{tab:appendix_table_rougelsum}
\resizebox{\textwidth}{!}{%
\begin{tabular}{l l *{4}{S} S *{3}{S} S}
\toprule
& & \multicolumn{5}{c}{\textbf{\texttt{DomainBench}}} & \multicolumn{4}{c}{\textbf{\texttt{InstructBench}}}\\
\cmidrule(lr){3-7} \cmidrule(lr){8-11}
\textbf{Model} & \textbf{Method} &
\texttt{Agriculture} & \texttt{GeoSignal} & \texttt{GenMedGPT} & \texttt{Wealth} &
\multicolumn{1}{c}{\textbf{Avg.}} &
\texttt{Dolly} & \texttt{Alpaca-GPT4} & \texttt{InstructWild} &
\multicolumn{1}{c}{\textbf{Avg.}} \\
\midrule
\multirow{12}{*}{\textsc{Llama-3.2-3B}}
& Base Model & 8.34 & 22.02 & 14.13 & 21.45 & 16.48 & 30.68 & 34.41 & 25.61 & 30.23 \\
& Tent & 0.98 & 4.59 & 9.32 & 2.33 & 4.30 & 5.66 & 5.72 & 6.41 & 5.93 \\
& EATA & 0.39 & 4.89 & 5.49 & 0.03 & 2.70 & 1.05 & 6.83 & 3.55 & 3.81 \\
& TLM & 14.23 & 27.56 & \bestscore{24.29} & 26.73 & 23.20 & 24.77 & 37.66 & 27.66 & 30.03 \\
& \multicolumn{10}{>{\columncolor{gray!10}}c}{\modewocache} \\
& \method{}-2 & 18.86 & \bestscore{28.99} & \secondscore{24.65} & \bestscore{29.14} & \bestscore{25.41} & 29.31 & 40.38 & 32.48 & 34.06 \\
& \method{}-4 & \secondscore{19.72} & 26.74 & 17.64 & \secondscore{29.10} & 23.30 & 32.56 & \bestscore{40.53} & \secondscore{34.69} & \secondscore{35.93} \\
& \method{}-8 & 18.56 & 28.70 & 20.10 & 28.53 & 23.97 & 32.69 & 40.00 & 33.26 & 35.32 \\
& \method{}-16 & 18.37 & 27.15 & 17.85 & 28.18 & 22.89 & 30.56 & \secondscore{39.72} & 32.08 & 34.12 \\
& \multicolumn{10}{>{\columncolor{gray!10}}c}{\modecache} \\
& \method{}-2 & 15.18 & 27.48 & 18.39 & 28.80 & 22.46 & \secondscore{34.12} & 39.88 & 32.23 & 35.49 \\
& \method{}-4 & \bestscore{20.12} & \secondscore{29.45} & 17.59 & 29.07 & \secondscore{24.06} & \bestscore{34.12} & \bestscore{40.53} & \bestscore{36.15} & \bestscore{36.93} \\
& \method{}-8 & 16.95 & 28.21 & 21.01 & 28.30 & 23.62 & 34.08 & 39.05 & 35.10 & 36.08 \\
& \method{}-16 & 15.38 & 28.31 & 19.66 & 28.28 & 22.91 & 33.46 & 39.67 & 32.65 & 35.26 \\

\midrule

\multirow{12}{*}{\textsc{Llama-3.1-8B}}
& Base Model & 8.59 & 22.28 & 13.53 & 21.65 & 16.51 & 32.90 & 34.40 & 25.67 & 30.99 \\
& Tent & 1.16 & 3.79 & 0.74 & 13.22 & 4.73 & 0.45 & 4.84 & 9.78 & 5.02 \\
& EATA & 1.52 & 6.43 & 1.86 & 14.60 & 6.10 & 1.75 & 5.89 & 2.53 & 3.39 \\
& TLM & 16.33 & 28.85 & 25.71 & 28.95 & 24.96 & 32.36 & 38.41 & 28.88 & 33.22 \\
& \multicolumn{10}{>{\columncolor{gray!10}}c}{\modewocache} \\
& \method{}-2 & 17.60 & \bestscore{30.38} & 25.83 & 29.51 & 25.83 & 33.39 & 40.84 & 30.52 & 34.92 \\
& \method{}-4 & \bestscore{20.17} & \secondscore{29.47} & \secondscore{26.48} & \secondscore{29.58} & \secondscore{26.43} & 34.61 & \bestscore{41.27} & \bestscore{36.15} & \bestscore{37.34} \\
& \method{}-8 & \secondscore{20.44} & 29.00 & \bestscore{26.66} & \bestscore{30.03} & \bestscore{26.53} & 32.23 & 40.60 & 34.34 & 35.72 \\
& \method{}-16 & 19.56 & 26.52 & 25.03 & 29.55 & 25.16 & 32.98 & 39.45 & 35.05 & 35.83 \\
& \multicolumn{10}{>{\columncolor{gray!10}}c}{\modecache} \\
& \method{}-2 & 16.40 & 29.04 & 21.97 & 29.83 & 24.31 & 33.47 & 40.54 & 30.67 & 34.90 \\
& \method{}-4 & 16.49 & \bestscore{29.52} & 24.82 & 29.50 & 25.08 & \bestscore{35.45} & \secondscore{40.85} & \secondscore{35.71} & \bestscore{37.34} \\
& \method{}-8 & 16.56 & 29.87 & 25.30 & 29.24 & 25.24 & 35.19 & 39.82 & 33.17 & 36.06 \\
& \method{}-16 & 15.17 & 29.19 & 21.23 & \secondscore{29.86} & 23.86 & \secondscore{35.42} & 39.85 & 32.08 & 35.78 \\

\midrule

\multirow{12}{*}{\textsc{Qwen-2.5-7B}}
& Base Model & 9.43 & 22.03 & 12.51 & 23.88 & 16.96 & 27.05 & 38.17 & 27.77 & 31.00 \\
& Tent & 19.64 & 22.15 & 5.31 & 28.59 & 18.92 & 21.47 & 24.38 & 26.93 & 24.26 \\
& EATA & 16.30 & 21.24 & 12.83 & 22.57 & 18.23 & 30.57 & 27.01 & 23.81 & 27.13 \\
& TLM & 11.23 & 26.21 & 29.67 & 28.13 & 23.81 & 31.05 & 43.08 & 30.76 & 34.96 \\
& \multicolumn{10}{>{\columncolor{gray!10}}c}{\modewocache} \\
& \method{}-2 & 15.22 & 29.55 & 28.96 & 29.12 & 25.71 & 36.07 & \secondscore{43.33} & 31.67 & 37.02 \\
& \method{}-4 & 17.68 & \secondscore{29.42} & \bestscore{29.74} & 29.69 & 26.63 & 35.69 & \secondscore{43.33} & 32.95 & 37.32 \\
& \method{}-8 & \secondscore{21.79} & 29.12 & 29.28 & \secondscore{29.93} & \secondscore{27.53} & 35.97 & 42.74 & \bestscore{34.35} & \bestscore{37.69} \\
& \method{}-16 & \bestscore{21.14} & 28.81 & 26.83 & \bestscore{30.25} & 26.76 & 35.93 & 43.13 & 33.67 & 37.58 \\
& \multicolumn{10}{>{\columncolor{gray!10}}c}{\modecache} \\
& \method{}-2 & 13.47 & 29.75 & 29.48 & 29.08 & 25.45 & 35.46 & 42.90 & 31.31 & 36.56 \\
& \method{}-4 & 19.40 & 29.37 & \secondscore{29.56} & 29.67 & \bestscore{27.00} & \bestscore{36.51} & \bestscore{43.40} & \bestscore{34.07} & \bestscore{37.99} \\
& \method{}-8 & 21.19 & \bestscore{29.58} & 28.90 & 29.82 & 27.37 & \secondscore{36.27} & 42.04 & 33.63 & 37.31 \\
& \method{}-16 & 18.31 & \secondscore{29.47} & 25.92 & 29.79 & 25.87 & \secondscore{36.27} & 43.04 & \secondscore{33.72} & \secondscore{37.68} \\

\midrule

\multirow{12}{*}{\textsc{Qwen-2.5-14B}}
& Base Model & 10.67 & 23.46 & 14.42 & 24.36 & 18.23 & 28.06 & 39.34 & 28.12 & 31.84 \\
& Tent & 4.92 & 27.89 & 14.87 & 28.19 & 18.97 & 29.66 & 28.12 & 11.29 & 23.02 \\
& EATA & 1.88 & 28.23 & 3.17 & 27.97 & 15.31 & 22.99 & 26.08 & 25.33 & 24.80 \\
& TLM & 11.09 & 28.70 & \bestscore{32.20} & 29.48 & 25.37 & 34.04 & 42.20 & 30.59 & 35.61 \\
& \multicolumn{10}{>{\columncolor{gray!10}}c}{\modewocache} \\
& \method{}-2 & 16.37 & 30.52 & \secondscore{30.45} & 29.88 & 26.80 & 35.60 & \bestscore{43.37} & 32.86 & 37.28 \\
& \method{}-4 & \secondscore{20.09} & 30.57 & \secondscore{31.05} & \bestscore{30.14} & \bestscore{27.96} & \bestscore{37.04} & \secondscore{43.24} & 34.45 & \bestscore{38.24} \\
& \method{}-8 & \bestscore{22.01} & \bestscore{31.31} & 24.51 & \secondscore{30.05} & 26.97 & 35.71 & 42.65 & \bestscore{36.56} & \secondscore{38.31} \\
& \method{}-16 & 18.82 & 28.72 & 29.79 & 29.95 & 26.82 & 35.57 & 42.86 & 35.49 & 37.97 \\
& \multicolumn{10}{>{\columncolor{gray!10}}c}{\modecache} \\
& \method{}-2 & 14.24 & 30.14 & 29.73 & 28.65 & 25.69 & \secondscore{36.88} & 43.17 & 31.58 & 37.21 \\
& \method{}-4 & 19.52 & 30.45 & 28.91 & 29.53 & \secondscore{27.10} & 36.32 & 43.13 & 34.12 & 37.86 \\
& \method{}-8 & 16.34 & 30.91 & 25.51 & 29.86 & 25.65 & 36.28 & 42.62 & 34.52 & 37.81 \\
& \method{}-16 & \secondscore{21.85} & \secondscore{30.93} & 22.26 & 29.57 & 26.15 & \bestscore{37.04} & 42.90 & \secondscore{34.46} & 38.13 \\

\bottomrule
\end{tabular}
}
\end{table*}

%% file: tex/appendix_table_bertscore.tex
\begin{table*}[t]
\small
\centering
\caption{\small{Detailed results on \texttt{DomainBench} and \texttt{InstructBench}. BERTScore-F1 scores ($\times 100$ for visibility; higher is better). For each model and dataset, the highest score is \textbf{bold} and the second-highest is \underline{underlined}.}}
\vspace{0.5em}
\label{tab:appendix_table_bertscoref1}
\resizebox{\textwidth}{!}{%
\begin{tabular}{l l *{4}{S} S *{3}{S} S}
\toprule
& & \multicolumn{5}{c}{\textbf{\texttt{DomainBench}}} & \multicolumn{4}{c}{\textbf{\texttt{InstructBench}}} \\
\cmidrule(lr){3-7} \cmidrule(lr){8-11}
\textbf{Model} & \textbf{Method} &
\texttt{Agriculture} & \texttt{GeoSignal} & \texttt{GenMedGPT} & \texttt{Wealth} &
\multicolumn{1}{c}{\textbf{Avg.}} &
\texttt{Dolly} & \texttt{Alpaca-GPT4} & \texttt{InstructWild} &
\multicolumn{1}{c}{\textbf{Avg.}} \\
\midrule
\multirow{14}{*}{\textsc{Llama-3.2-3B}}
& Base Model & 66.66 & 67.72 & 66.74 & 67.75 & 67.22 & 71.74 & 72.11 & 70.18 & 71.34 \\
& Tent & 66.62 & 64.49 & 66.78 & 64.43 & 65.58 & 68.04 & 72.82 & 69.75 & 70.20 \\
& EATA & 67.44 & 69.45 & 68.01 & 63.21 & 67.03 & 67.80 & 71.92 & 66.50 & 68.74 \\
& TLM & 66.28 & \bestscore{70.17} & \bestscore{70.95} & 69.30 & 69.17 & 69.87 & 74.32 & 70.61 & 71.60 \\
& \multicolumn{10}{>{\columncolor{gray!10}}c}{\modewocache} \\
& \method{}-2 & 69.25 & 70.06 & \secondscore{70.77} & 70.05 & \bestscore{70.03} & 71.91 & \secondscore{74.70} & 72.03 & 72.88 \\
& \method{}-4 & \secondscore{69.94} & 69.50 & 69.24 & \secondscore{70.27} & 69.74 & 72.75 & \bestscore{74.71} & 72.06 & 73.17 \\
& \method{}-8 & 68.98 & 69.96 & 70.13 & 70.22 & \secondscore{69.82} & 72.59 & 74.56 & 71.41 & 72.86 \\
& \method{}-16 & 69.68 & 68.27 & 69.79 & 69.70 & 69.36 & 69.96 & 72.88 & 71.42 & 71.42 \\
& \multicolumn{10}{>{\columncolor{gray!10}}c}{\modecache} \\
& \method{}-2 & 66.49 & 70.03 & 69.09 & 70.02 & 68.91 & \bestscore{73.51} & 74.37 & 71.41 & 73.10 \\
& \method{}-4 & \bestscore{70.01} & \secondscore{70.16} & 68.54 & \bestscore{70.28} & 69.75 & 73.11 & 74.70 & \bestscore{72.87} & \bestscore{73.56} \\
& \method{}-8 & 68.05 & 69.90 & 69.99 & 70.03 & 69.49 & \secondscore{73.35} & 74.15 & \secondscore{72.22} & \secondscore{73.24} \\
& \method{}-16 & 66.42 & 69.32 & 69.80 & 69.72 & 68.81 & 73.29 & 72.85 & 71.42 & 72.52 \\
\midrule
\multirow{14}{*}{\textsc{Llama-3.1-8B}}
& Base Model & 66.66 & 67.77 & 66.41 & 67.72 & 67.14 & 72.91 & 72.05 & 70.14 & 71.70 \\
& Tent & 67.46 & 69.03 & 67.69 & 67.91 & 68.02 & 67.76 & 67.87 & 69.12 & 68.25 \\
& EATA & 66.28 & 67.34 & 66.46 & 65.45 & 66.38 & 73.83 & 59.81 & 68.98 & 67.54 \\
& TLM & 66.89 & 70.86 & \secondscore{72.10} & 69.91 & 69.94 & 71.89 & 74.44 & 70.80 & 72.37 \\
& \multicolumn{10}{>{\columncolor{gray!10}}c}{\modewocache} \\
& \method{}-2 & 67.26 & 70.89 & 71.44 & 69.95 & 69.89 & 73.24 & \bestscore{74.91} & 71.27 & 73.14 \\
& \method{}-4 & \secondscore{70.29} & 70.82 & \bestscore{72.78} & 69.85 & \bestscore{70.94} & 73.83 & 74.73 & 72.45 & \secondscore{73.67} \\
& \method{}-8 & \bestscore{70.35} & \secondscore{70.95} & 71.73 & \bestscore{70.22} & \secondscore{70.82} & 73.00 & \secondscore{74.76} & 72.03 & 73.26 \\
& \method{}-16 & 70.19 & 69.03 & 70.92 & \secondscore{70.09} & 70.06 & 71.01 & 74.08 & \bestscore{74.00} & 73.03 \\
& \multicolumn{10}{>{\columncolor{gray!10}}c}{\modecache} \\
& \method{}-2 & 67.00 & \bestscore{71.02} & 69.72 & 69.80 & 69.38 & 73.32 & 74.52 & 71.30 & 73.05 \\
& \method{}-4 & 67.28 & 70.32 & 71.29 & 69.88 & 69.69 & \bestscore{73.99} & 74.48 & \secondscore{72.94} & \bestscore{73.81} \\
& \method{}-8 & 67.22 & 70.75 & 71.78 & 69.95 & 69.93 & \secondscore{73.95} & 74.37 & 71.45 & 73.26 \\
& \method{}-16 & 66.83 & 70.30 & 69.97 & 69.72 & 69.21 & 73.74 & 74.37 & 72.24 & 73.45 \\
\midrule
\multirow{14}{*}{\textsc{Qwen-2.5-7B}}
& Base Model & 65.67 & 67.91 & 65.51 & 68.43 & 66.88 & 70.60 & 73.56 & 70.61 & 71.59 \\
& Tent & 69.06 & 70.40 & 67.00 & 68.87 & 68.83 & 70.54 & 74.42 & 70.78 & 71.91 \\
& EATA & 66.34 & 69.97 & 67.05 & 68.62 & 68.00 & 71.17 & 72.92 & 71.14 & 71.74 \\
& TLM & 64.99 & 70.07 & \bestscore{74.07} & 70.22 & 69.84 & 73.15 & \bestscore{75.95} & 71.65 & 73.58 \\
& \multicolumn{10}{>{\columncolor{gray!10}}c}{\modewocache} \\
& \method{}-2 & 66.54 & 71.02 & 73.42 & 69.96 & 70.24 & 73.54 & \secondscore{75.94} & 71.58 & 73.68 \\
& \method{}-4 & 69.78 & 71.03 & 73.48 & 70.26 & 71.14 & 74.33 & 75.75 & 71.77 & 73.95 \\
& \method{}-8 & \bestscore{71.07} & \bestscore{71.24} & 73.66 & \secondscore{70.70} & \bestscore{71.67} & 74.44 & 75.71 & \bestscore{72.53} & 74.23 \\
& \method{}-16 & \secondscore{70.20} & 70.81 & 72.30 & \bestscore{70.89} & 71.05 & 74.10 & 75.30 & 71.87 & 73.76 \\
& \multicolumn{10}{>{\columncolor{gray!10}}c}{\modecache} \\
& \method{}-2 & 65.86 & \secondscore{71.21} & \secondscore{73.69} & 69.94 & 70.17 & 74.04 & 75.83 & 71.64 & 73.84 \\
& \method{}-4 & 68.90 & 70.73 & 73.56 & 70.05 & 70.81 & \secondscore{74.98} & 75.57 & 72.14 & \secondscore{74.23} \\
& \method{}-8 & 70.19 & 71.03 & 73.35 & 70.39 & \secondscore{71.24} & \bestscore{75.27} & 75.40 & 72.14 & \bestscore{74.27} \\
& \method{}-16 & 69.09 & 70.73 & 72.06 & 70.60 & 70.62 & 74.55 & 75.60 & \secondscore{72.43} & 74.19 \\
\midrule
\multirow{14}{*}{\textsc{Qwen-2.5-14B}}
& Base Model & 65.21 & 68.28 & 65.98 & 68.33 & 66.95 & 70.63 & 73.87 & 70.91 & 71.80 \\
& Tent & 68.01 & 69.39 & 68.42 & 69.92 & 68.94 & 73.72 & 74.25 & 69.80 & 72.59 \\
& EATA & 64.73 & 70.27 & 68.02 & 69.00 & 68.00 & 73.98 & 74.42 & 70.95 & 73.11 \\
& TLM & 64.81 & 70.83 & \bestscore{75.38} & 70.46 & 70.37 & 73.34 & 76.13 & 71.58 & 73.68 \\
& \multicolumn{10}{>{\columncolor{gray!10}}c}{\modewocache} \\
& \method{}-2 & 66.22 & 71.36 & 74.31 & 70.08 & 70.50 & 73.68 & \secondscore{76.25} & 71.85 & 73.93 \\
& \method{}-4 & 68.35 & 71.06 & \secondscore{74.48} & 70.29 & \secondscore{71.04} & 73.72 & 76.09 & 72.06 & 73.96 \\
& \method{}-8 & \secondscore{71.05} & \bestscore{71.97} & 72.04 & \bestscore{70.70} & \bestscore{71.44} & 73.50 & 75.93 & \bestscore{74.49} & \bestscore{74.64} \\
& \method{}-16 & 68.86 & 70.69 & 73.91 & \secondscore{70.56} & 71.00 & 73.98 & 75.46 & \secondscore{72.31} & 73.92 \\
& \multicolumn{10}{>{\columncolor{gray!10}}c}{\modecache} \\
& \method{}-2 & 65.40 & 71.04 & 73.92 & 69.76 & 70.03 & 74.33 & \bestscore{76.26} & 71.53 & 74.04 \\
& \method{}-4 & 68.38 & 71.33 & 73.71 & 69.99 & 70.85 & 74.22 & 76.05 & 72.03 & 74.10 \\
& \method{}-8 & 65.99 & 71.20 & 72.35 & 70.54 & 70.02 & \bestscore{74.61} & 75.90 & 71.98 & \secondscore{74.17} \\
& \method{}-16 & \bestscore{71.37} & \secondscore{71.72} & 70.87 & 70.10 & 71.01 & \secondscore{74.56} & 75.85 & 71.96 & 74.12 \\
\bottomrule
\end{tabular}
}
\end{table*}

%% file: tex/appendix_table_rouge1.tex
\begin{table*}[t]
\small
\centering
\caption{\small{Detailed results on \texttt{DomainBench} and \texttt{InstructBench}. ROUGE-1 scores ($\times 100$ for visibility; higher is better). For each model and dataset, the highest score is \textbf{bold} and the second-highest is \underline{underlined}.}}
\vspace{0.5em}
\label{tab:appendix_table_rouge1}
\resizebox{\textwidth}{!}{%
\begin{tabular}{l l *{4}{S} S *{3}{S} S}
\toprule
& & \multicolumn{5}{c}{\textbf{\texttt{DomainBench}}} & \multicolumn{4}{c}{\textbf{\texttt{InstructBench}}} \\
\cmidrule(lr){3-7} \cmidrule(lr){8-11}
\textbf{Model} & \textbf{Method} &
\texttt{Agriculture} & \texttt{GeoSignal} & \texttt{GenMedGPT} & \texttt{Wealth} &
\multicolumn{1}{c}{\textbf{Avg.}} &
\texttt{Dolly} & \texttt{Alpaca-GPT4} & \texttt{InstructWild} &
\multicolumn{1}{c}{\textbf{Avg.}} \\
\midrule
\multirow{14}{*}{\textsc{Llama-3.2-3B}}
& Base Model & 9.09 & 24.57 & 15.89 & 23.39 & 18.24 & 34.66 & 37.60 & 27.97 & 33.41 \\
& Tent & 9.02 & 20.41 & 17.71 & 16.96 & 16.03 & 28.73 & 40.81 & 26.56 & 32.04 \\
& EATA & 9.45 & 29.32 & 16.55 & 18.19 & 18.37 & 25.89 & 36.48 & 21.91 & 28.10 \\
& TLM & 16.11 & 30.98 & \secondscore{26.76} & 29.35 & 25.80 & 28.20 & 40.97 & 30.30 & 33.16 \\
& \multicolumn{10}{>{\columncolor{gray!10}}c}{\modewocache} \\
& \method{}-2 & 21.91 & \bestscore{32.45} & \bestscore{27.00} & \bestscore{32.03} & \bestscore{28.35} & 33.30 & 44.03 & 35.66 & 37.67 \\
& \method{}-4 & \secondscore{22.76} & 29.75 & 18.47 & \secondscore{31.86} & 25.71 & 36.24 & \secondscore{44.05} & 38.34 & 39.55 \\
& \method{}-8 & 21.58 & 32.11 & 22.57 & 31.55 & \secondscore{26.95} & 37.39 & 43.40 & 36.64 & 39.14 \\
& \method{}-16 & 21.48 & 23.05 & 20.96 & 30.73 & 24.05 & 27.76 & 40.81 & 35.32 & 34.63 \\
& \multicolumn{10}{>{\columncolor{gray!10}}c}{\modecache} \\
& \method{}-2 & 17.32 & 30.92 & 21.83 & 31.78 & 25.46 & \bestscore{38.83} & 43.48 & 35.48 & 39.26 \\
& \method{}-4 & \bestscore{23.65} & \secondscore{32.18} & 17.36 & 31.84 & 26.26 & 37.60 & \bestscore{44.08} & \bestscore{39.83} & \bestscore{40.51} \\
& \method{}-8 & 19.59 & 31.30 & 23.36 & 31.08 & 26.33 & \secondscore{38.40} & 42.49 & \secondscore{38.72} & \secondscore{39.87} \\
& \method{}-16 & 17.64 & 29.58 & 20.98 & 30.35 & 24.64 & 37.76 & 39.13 & 35.31 & 37.40 \\
\midrule
\multirow{14}{*}{\textsc{Llama-3.1-8B}}
& Base Model & 9.36 & 24.92 & 15.07 & 23.61 & 18.24 & 36.85 & 37.56 & 28.02 & 34.14 \\
& Tent & 11.41 & 28.54 & 17.48 & 25.51 & 20.73 & 27.44 & 28.21 & 27.19 & 27.61 \\
& EATA & 9.66 & 27.71 & 15.97 & 13.89 & 16.81 & 38.52 & 7.77 & 29.88 & 25.39 \\
& TLM & 18.35 & 32.27 & 28.17 & 31.71 & 27.63 & 36.44 & 41.72 & 31.73 & 36.63 \\
& \multicolumn{10}{>{\columncolor{gray!10}}c}{\modewocache} \\
& \method{}-2 & 20.07 & \bestscore{33.85} & 28.23 & 32.37 & 28.63 & 37.69 & \secondscore{44.44} & 33.57 & 38.57 \\
& \method{}-4 & \secondscore{23.08} & 32.90 & \bestscore{28.78} & 32.33 & \secondscore{29.27} & 38.52 & \bestscore{44.67} & \bestscore{40.09} & \secondscore{41.09} \\
& \method{}-8 & \bestscore{23.97} & 32.29 & \secondscore{28.58} & \bestscore{32.77} & \bestscore{29.40} & 36.35 & 44.14 & 37.81 & 39.43 \\
& \method{}-16 & 22.70 & 28.54 & 27.08 & 30.91 & 27.31 & 28.34 & 41.40 & 38.94 & 36.23 \\
& \multicolumn{10}{>{\columncolor{gray!10}}c}{\modecache} \\
& \method{}-2 & 18.57 & \secondscore{33.53} & 24.44 & \secondscore{32.54} & 27.27 & 38.10 & 44.31 & 33.70 & 38.70 \\
& \method{}-4 & 16.87 & 32.85 & 27.38 & 32.26 & 27.34 & 39.80 & 44.12 & \secondscore{39.50} & \bestscore{41.14} \\
& \method{}-8 & 18.74 & 33.37 & 27.35 & 32.06 & 27.88 & \secondscore{40.25} & 43.45 & 36.60 & 40.10 \\
& \method{}-16 & 16.93 & 32.31 & 24.80 & 32.09 & 26.53 & \bestscore{40.33} & 42.80 & 35.08 & 39.40 \\
\midrule
\multirow{14}{*}{\textsc{Qwen-2.5-7B}}
& Base Model & 10.18 & 24.48 & 13.61 & 25.98 & 18.56 & 30.10 & 41.52 & 30.15 & 33.92 \\
& Tent & 22.45 & 32.45 & 17.26 & 31.28 & 25.86 & 32.15 & 43.77 & 33.42 & 36.45 \\
& EATA & 16.31 & 31.91 & 16.76 & 29.61 & 23.65 & 32.72 & 40.90 & 35.46 & 36.36 \\
& TLM & 11.96 & 29.98 & \bestscore{32.76} & 31.72 & 26.60 & 37.43 & 44.60 & 34.65 & 38.89 \\
& \multicolumn{10}{>{\columncolor{gray!10}}c}{\modewocache} \\
& \method{}-2 & 16.98 & 33.17 & 31.41 & 31.76 & 28.33 & 40.76 & \bestscore{46.65} & 34.73 & 40.71 \\
& \method{}-4 & 23.94 & 33.02 & 31.97 & 32.41 & 30.33 & 40.75 & 45.05 & 36.22 & 40.67 \\
& \method{}-8 & \bestscore{25.99} & \secondscore{33.22} & 31.64 & \secondscore{32.71} & \bestscore{30.89} & 41.35 & 46.24 & \bestscore{37.99} & \bestscore{41.86} \\
& \method{}-16 & \secondscore{24.79} & 32.23 & 29.48 & \bestscore{33.12} & 29.90 & 40.83 & 45.84 & 36.66 & 41.11 \\
& \multicolumn{10}{>{\columncolor{gray!10}}c}{\modecache} \\
& \method{}-2 & 14.87 & \bestscore{33.26} & 32.01 & 31.78 & 27.98 & 40.37 & \secondscore{46.42} & 34.35 & 40.38 \\
& \method{}-4 & 22.45 & 32.74 & \secondscore{32.03} & 32.33 & 29.89 & \secondscore{41.66} & 45.33 & \secondscore{37.59} & 41.52 \\
& \method{}-8 & 24.72 & 33.11 & 31.33 & 32.57 & \secondscore{30.43} & \bestscore{42.08} & 45.61 & 37.02 & \secondscore{41.57} \\
& \method{}-16 & 20.85 & 32.96 & 28.61 & 32.53 & 28.74 & 40.50 & 46.15 & 37.10 & 41.25 \\
\midrule
\multirow{14}{*}{\textsc{Qwen-2.5-14B}}
& Base Model & 11.67 & 26.09 & 15.99 & 26.59 & 20.09 & 31.16 & 42.76 & 30.58 & 34.83 \\
& Tent & 16.99 & 31.07 & 20.08 & 30.96 & 24.78 & 40.09 & 42.74 & 32.26 & 38.36 \\
& EATA & 13.41 & 30.59 & 21.03 & 29.93 & 23.74 & 40.26 & 43.81 & 32.09 & 38.72 \\
& TLM & 12.12 & 32.01 & \bestscore{34.53} & 32.31 & 27.74 & 37.94 & 45.64 & 33.33 & 38.97 \\
& \multicolumn{10}{>{\columncolor{gray!10}}c}{\modewocache} \\
& \method{}-2 & 18.33 & 34.17 & \secondscore{32.94} & 32.70 & 29.53 & 40.38 & \bestscore{46.87} & 36.04 & 41.10 \\
& \method{}-4 & 23.26 & 33.38 & 32.91 & 32.43 & \bestscore{30.50} & 40.87 & 46.69 & \secondscore{37.98} & 41.85 \\
& \method{}-8 & \bestscore{26.00} & \bestscore{34.80} & 27.35 & \bestscore{33.09} & \secondscore{30.31} & 40.15 & 46.28 & \bestscore{40.65} & \bestscore{42.36} \\
& \method{}-16 & 21.43 & 32.07 & 32.16 & \secondscore{32.89} & 29.64 & 40.26 & 45.17 & 37.94 & 41.13 \\
& \multicolumn{10}{>{\columncolor{gray!10}}c}{\modecache} \\
& \method{}-2 & 15.91 & 33.67 & 32.27 & 31.29 & 28.28 & \secondscore{41.68} & \secondscore{46.73} & 34.60 & 41.00 \\
& \method{}-4 & 22.61 & 33.91 & 31.41 & 32.42 & 30.09 & 41.55 & 46.63 & 37.45 & \secondscore{41.88} \\
& \method{}-8 & 18.41 & 34.30 & 28.55 & 32.75 & 28.50 & 41.00 & 46.20 & 37.88 & 41.69 \\
& \method{}-16 & \secondscore{25.57} & \secondscore{34.45} & 25.38 & 32.47 & 29.47 & \bestscore{41.81} & 45.75 & 36.87 & 41.47 \\
\bottomrule
\end{tabular}
}
\end{table*}

%% file: tex/appendix_table_rouge2.tex
\begin{table*}[t]
\small
\centering
\caption{\small{Detailed results on \texttt{DomainBench} and \texttt{InstructBench}. ROUGE-2 scores ($\times 100$ for visibility; higher is better). For each model and dataset, the highest score is \textbf{bold} and the second-highest is \underline{underlined}.}}
\vspace{0.5em}
\label{tab:appendix_table_rouge2}
\resizebox{\textwidth}{!}{%
\begin{tabular}{l l *{4}{S} S *{3}{S} S}
\toprule
& & \multicolumn{5}{c}{\textbf{\texttt{DomainBench}}} & \multicolumn{4}{c}{\textbf{\texttt{InstructBench}}} \\
\cmidrule(lr){3-7} \cmidrule(lr){8-11}
\textbf{Model} & \textbf{Method} &
\texttt{Agriculture} & \texttt{GeoSignal} & \texttt{GenMedGPT} & \texttt{Wealth} &
\multicolumn{1}{c}{\textbf{Avg.}} &
\texttt{Dolly} & \texttt{Alpaca-GPT4} & \texttt{InstructWild} &
\multicolumn{1}{c}{\textbf{Avg.}} \\
\midrule
\multirow{14}{*}{\textsc{Llama-3.2-3B}}
& Base Model & 3.04 & 9.81 & 2.70 & 7.42 & 5.74 & 16.56 & 16.12 & 9.39 & 14.03 \\
& Tent & 3.05 & 8.21 & 2.77 & 5.17 & 4.80 & 13.68 & 18.13 & 9.41 & 13.74 \\
& EATA & 3.11 & 12.96 & 3.18 & 7.21 & 6.61 & 12.08 & 15.60 & 7.61 & 11.76 \\
& TLM & 5.37 & 14.89 & \secondscore{8.49} & 10.58 & 9.83 & 12.30 & 19.02 & 10.90 & 14.08 \\
& \multicolumn{10}{>{\columncolor{gray!10}}c}{\modewocache} \\
& \method{}-2 & \secondscore{7.07} & \secondscore{15.05} & \bestscore{9.21} & 11.91 & \bestscore{10.81} & 15.41 & \bestscore{20.59} & 14.16 & 16.72 \\
& \method{}-4 & 6.57 & 13.76 & 4.12 & \bestscore{11.95} & 9.10 & 17.59 & 20.43 & \secondscore{14.72} & 17.58 \\
& \method{}-8 & 6.60 & \bestscore{15.18} & 6.66 & 11.57 & \secondscore{10.00} & 18.19 & 20.10 & 13.78 & 17.35 \\
& \method{}-16 & 6.25 & 9.49 & 3.64 & 10.63 & 7.50 & 12.70 & 18.13 & 12.81 & 14.55 \\
& \multicolumn{10}{>{\columncolor{gray!10}}c}{\modecache} \\
& \method{}-2 & 5.73 & 13.89 & 4.19 & 11.62 & 8.86 & \bestscore{19.52} & 20.02 & 12.85 & 17.46 \\
& \method{}-4 & \bestscore{7.11} & 14.83 & 3.76 & \secondscore{11.92} & 9.40 & \secondscore{18.99} & \secondscore{20.47} & \bestscore{15.78} & \bestscore{18.41} \\
& \method{}-8 & 6.03 & 14.41 & 7.37 & 11.24 & 9.76 & 18.71 & 19.56 & 14.57 & \secondscore{17.61} \\
& \method{}-16 & 5.07 & 13.39 & 3.62 & 10.63 & 8.18 & 18.58 & 17.13 & 12.79 & 16.17 \\
\midrule
\multirow{14}{*}{\textsc{Llama-3.1-8B}}
& Base Model & 3.32 & 9.97 & 3.28 & 7.48 & 6.01 & 18.40 & 16.24 & 9.28 & 14.64 \\
& Tent & 2.62 & 11.82 & 2.62 & 8.51 & 6.39 & 12.81 & 10.69 & 9.65 & 11.05 \\
& EATA & 3.32 & 12.33 & 2.22 & 2.94 & 5.20 & 18.92 & 2.41 & 10.29 & 10.54 \\
& TLM & 6.43 & 15.17 & 9.61 & 11.89 & 10.78 & 18.21 & 20.02 & 11.43 & 16.56 \\
& \multicolumn{10}{>{\columncolor{gray!10}}c}{\modewocache} \\
& \method{}-2 & 6.41 & \bestscore{16.47} & 10.99 & \secondscore{12.27} & 11.54 & 18.81 & \bestscore{21.14} & 13.93 & 17.96 \\
& \method{}-4 & 7.05 & 15.82 & \secondscore{11.14} & 12.17 & \secondscore{11.54} & 18.92 & \secondscore{21.00} & 14.83 & \secondscore{18.25} \\
& \method{}-8 & \bestscore{7.26} & 15.31 & \bestscore{13.22} & \bestscore{12.48} & \bestscore{12.07} & 17.24 & 20.57 & 14.26 & 17.36 \\
& \method{}-16 & \secondscore{7.26} & 11.82 & 9.36 & 10.75 & 9.80 & 13.44 & 19.11 & \secondscore{14.98} & 15.84 \\
& \multicolumn{10}{>{\columncolor{gray!10}}c}{\modecache} \\
& \method{}-2 & 6.08 & 15.64 & 7.33 & 12.11 & 10.29 & 18.99 & 20.60 & 11.94 & 17.18 \\
& \method{}-4 & 5.73 & \secondscore{16.05} & 9.03 & 11.99 & 10.70 & \bestscore{20.73} & 20.72 & \bestscore{15.44} & \bestscore{18.96} \\
& \method{}-8 & 6.09 & 15.43 & 10.92 & 11.91 & 11.09 & 19.87 & 19.92 & 13.60 & 17.79 \\
& \method{}-16 & 5.65 & 14.61 & 7.23 & 11.71 & 9.80 & \secondscore{20.71} & 19.52 & 12.96 & 17.73 \\
\midrule
\multirow{14}{*}{\textsc{Qwen-2.5-7B}}
& Base Model & 3.63 & 9.50 & 3.50 & 8.67 & 6.33 & 14.33 & 18.64 & 10.11 & 14.36 \\
& Tent & 7.32 & 15.15 & 4.24 & 11.96 & 9.67 & 16.17 & 21.51 & 12.13 & 16.60 \\
& EATA & 5.32 & 15.09 & 4.10 & 11.60 & 9.03 & 16.50 & 19.68 & 13.21 & 16.46 \\
& TLM & 4.20 & 13.23 & \bestscore{14.85} & 11.79 & 11.02 & 19.27 & 22.34 & 12.60 & 18.07 \\
& \multicolumn{10}{>{\columncolor{gray!10}}c}{\modewocache} \\
& \method{}-2 & 6.07 & 15.85 & 13.67 & 12.04 & 11.91 & 21.93 & \bestscore{23.12} & 12.78 & 19.28 \\
& \method{}-4 & 7.71 & 15.83 & 14.10 & 12.64 & \secondscore{12.57} & 21.55 & 22.95 & 13.40 & 19.30 \\
& \method{}-8 & \bestscore{8.57} & \secondscore{15.85} & 14.18 & \bestscore{13.21} & \bestscore{12.95} & 22.29 & \secondscore{23.00} & \bestscore{14.86} & \bestscore{20.05} \\
& \method{}-16 & 8.05 & 15.28 & 11.38 & \secondscore{13.02} & 11.93 & 21.61 & 22.19 & 14.00 & 19.27 \\
& \multicolumn{10}{>{\columncolor{gray!10}}c}{\modecache} \\
& \method{}-2 & 5.31 & \bestscore{15.87} & \secondscore{14.27} & 11.83 & 11.82 & 21.33 & 22.89 & 12.54 & 18.92 \\
& \method{}-4 & 7.29 & 15.54 & 14.01 & 12.21 & 12.26 & \secondscore{22.53} & 22.79 & \secondscore{14.29} & \secondscore{19.87} \\
& \method{}-8 & \secondscore{8.10} & 15.62 & 13.61 & 12.79 & 12.53 & \bestscore{22.67} & 22.54 & 14.03 & 19.75 \\
& \method{}-16 & 6.86 & 15.84 & 10.86 & 12.54 & 11.53 & 21.83 & 22.21 & 14.15 & 19.39 \\
\midrule
\multirow{14}{*}{\textsc{Qwen-2.5-14B}}
& Base Model & 4.14 & 10.10 & 3.84 & 8.69 & 6.69 & 14.71 & 19.56 & 10.30 & 14.86 \\
& Tent & 5.64 & 14.21 & 4.42 & 11.20 & 8.87 & 21.38 & 20.17 & 11.03 & 17.53 \\
& EATA & 4.54 & 14.71 & 4.64 & 10.31 & 8.55 & 21.59 & 20.71 & 11.34 & 17.88 \\
& TLM & 4.31 & 14.40 & \bestscore{16.75} & 12.07 & 11.88 & 19.53 & 23.08 & 12.06 & 18.22 \\
& \multicolumn{10}{>{\columncolor{gray!10}}c}{\modewocache} \\
& \method{}-2 & 6.38 & \bestscore{16.53} & 14.76 & 12.35 & \secondscore{12.50} & 21.63 & \bestscore{23.38} & 13.21 & 19.41 \\
& \method{}-4 & 7.68 & 16.25 & 14.79 & 12.01 & \bestscore{12.68} & 20.66 & \secondscore{23.36} & 13.97 & 19.33 \\
& \method{}-8 & \bestscore{8.38} & \secondscore{16.36} & 9.43 & \bestscore{12.58} & 11.69 & 21.05 & 23.09 & \bestscore{16.29} & \bestscore{20.14} \\
& \method{}-16 & 6.98 & 15.31 & \secondscore{14.98} & 12.39 & 12.42 & 21.59 & 22.84 & \secondscore{14.27} & 19.56 \\
& \multicolumn{10}{>{\columncolor{gray!10}}c}{\modecache} \\
& \method{}-2 & 5.46 & 15.78 & 14.06 & 11.45 & 11.69 & 21.77 & 23.21 & 12.42 & 19.13 \\
& \method{}-4 & 7.52 & 15.79 & 13.29 & 12.04 & 12.16 & \bestscore{22.48} & 23.12 & 13.90 & \secondscore{19.83} \\
& \method{}-8 & 6.02 & 16.22 & 9.65 & \secondscore{12.51} & 11.10 & 22.22 & 23.12 & 14.09 & 19.81 \\
& \method{}-16 & \secondscore{8.19} & 15.37 & 7.19 & 12.01 & 10.69 & \secondscore{22.22} & 22.39 & 13.66 & 19.42 \\
\bottomrule
\end{tabular}
}
\end{table*}

%% file: tex/appendix_table_rougel.tex
\begin{table*}[t]
\small
\centering
\caption{\small{Detailed results on \texttt{DomainBench} and \texttt{InstructBench}. ROUGE-L scores ($\times 100$ for visibility; higher is better). For each model and dataset, the highest score is \textbf{bold} and the second-highest is \underline{underlined}.}}
\vspace{0.5em}
\label{tab:appendix_table_rougel}
\resizebox{\textwidth}{!}{%
\begin{tabular}{l l *{4}{S} S *{3}{S} S}
\toprule
& & \multicolumn{5}{c}{\textbf{\texttt{DomainBench}}} & \multicolumn{4}{c}{\textbf{\texttt{InstructBench}}} \\
\cmidrule(lr){3-7} \cmidrule(lr){8-11}
\textbf{Model} & \textbf{Method} &
\texttt{Agriculture} & \texttt{GeoSignal} & \texttt{GenMedGPT} & \texttt{Wealth} &
\multicolumn{1}{c}{\textbf{Avg.}} &
\texttt{Dolly} & \texttt{Alpaca-GPT4} & \texttt{InstructWild} &
\multicolumn{1}{c}{\textbf{Avg.}} \\
\midrule
\multirow{14}{*}{\textsc{Llama-3.2-3B}}
& Base Model & 6.63 & 17.20 & 10.37 & 15.16 & 12.34 & 26.02 & 24.93 & 16.97 & 22.64 \\
& Tent & 6.64 & 15.18 & 11.26 & 11.10 & 11.04 & 22.64 & 28.63 & 16.81 & 22.69 \\
& EATA & 7.00 & 21.78 & 11.61 & 15.40 & 13.95 & 20.20 & 24.25 & 15.20 & 19.88 \\
& TLM & 12.22 & 23.48 & \secondscore{19.68} & 20.30 & 18.92 & 20.88 & 29.59 & 19.16 & 23.21 \\
& \multicolumn{10}{>{\columncolor{gray!10}}c}{\modewocache} \\
& \method{}-2 & 16.82 & \secondscore{24.84} & \bestscore{19.91} & 22.37 & \bestscore{20.99} & 24.63 & \secondscore{31.06} & 24.08 & 26.59 \\
& \method{}-4 & \secondscore{17.48} & 22.19 & 13.58 & \secondscore{22.39} & 18.91 & 26.95 & 31.02 & \secondscore{25.70} & 27.89 \\
& \method{}-8 & 16.35 & \bestscore{24.96} & 16.25 & 22.35 & \secondscore{19.98} & 28.66 & 30.88 & 23.83 & 27.79 \\
& \method{}-16 & 16.07 & 17.13 & 13.14 & 21.89 & 17.06 & 21.29 & 28.63 & 22.49 & 24.14 \\
& \multicolumn{10}{>{\columncolor{gray!10}}c}{\modecache} \\
& \method{}-2 & 13.07 & 23.16 & 13.74 & 22.02 & 18.00 & \bestscore{30.27} & 30.56 & 22.09 & 27.64 \\
& \method{}-4 & \bestscore{18.51} & 24.41 & 13.22 & \bestscore{22.40} & 19.64 & 29.22 & \bestscore{31.07} & \bestscore{26.97} & \bestscore{29.09} \\
& \method{}-8 & 14.59 & 23.81 & 16.99 & 21.61 & 19.25 & \secondscore{29.84} & 30.17 & 24.96 & \secondscore{28.32} \\
& \method{}-16 & 13.34 & 22.86 & 13.14 & 20.88 & 17.56 & 29.39 & 27.81 & 22.48 & 26.56 \\
\midrule
\multirow{14}{*}{\textsc{Llama-3.1-8B}}
& Base Model & 6.84 & 17.38 & 9.87 & 15.23 & 12.33 & 28.07 & 24.86 & 16.94 & 23.29 \\
& Tent & 10.18 & 21.11 & 11.79 & 16.68 & 14.94 & 21.41 & 18.00 & 18.64 & 19.35 \\
& EATA & 7.01 & 21.35 & 10.91 & 8.73 & 12.00 & 30.31 & 7.55 & 18.72 & 18.86 \\
& TLM & 14.00 & 24.72 & 21.51 & 22.25 & 20.62 & 28.47 & 30.73 & 19.81 & 26.34 \\
& \multicolumn{10}{>{\columncolor{gray!10}}c}{\modewocache} \\
& \method{}-2 & 16.24 & \bestscore{26.72} & 22.49 & \secondscore{23.35} & 22.20 & 28.52 & \secondscore{31.74} & 22.59 & 27.62 \\
& \method{}-4 & \secondscore{17.34} & \secondscore{26.33} & \secondscore{22.95} & 22.51 & \secondscore{22.28} & 30.31 & \bestscore{32.00} & 25.22 & \secondscore{29.18} \\
& \method{}-8 & \bestscore{18.21} & 25.33 & \bestscore{23.99} & \bestscore{23.49} & \bestscore{22.76} & 27.65 & 31.11 & 24.75 & 27.84 \\
& \method{}-16 & 17.26 & 21.11 & 21.79 & 21.19 & 20.34 & 21.62 & 29.87 & \bestscore{26.31} & 25.94 \\
& \multicolumn{10}{>{\columncolor{gray!10}}c}{\modecache} \\
& \method{}-2 & 13.97 & 26.22 & 16.45 & 22.68 & 19.83 & 28.91 & 31.22 & 20.66 & 26.93 \\
& \method{}-4 & 12.60 & 25.86 & 19.39 & 22.11 & 19.99 & \bestscore{32.15} & 31.58 & \secondscore{26.29} & \bestscore{30.00} \\
& \method{}-8 & 14.00 & 25.22 & 19.71 & 22.50 & 20.36 & 30.68 & 30.62 & 23.16 & 28.16 \\
& \method{}-16 & 12.81 & 25.15 & 16.77 & 22.10 & 19.21 & \secondscore{31.25} & 30.30 & 23.07 & 28.21 \\
\midrule
\multirow{14}{*}{\textsc{Qwen-2.5-7B}}
& Base Model & 7.31 & 16.79 & 8.56 & 16.73 & 12.35 & 22.09 & 27.73 & 17.83 & 22.55 \\
& Tent & 17.85 & 25.05 & 10.70 & 22.23 & 18.96 & 25.24 & 32.28 & 20.83 & 26.12 \\
& EATA & 14.22 & 24.45 & 10.39 & 21.23 & 17.57 & 25.39 & 30.04 & 22.70 & 26.04 \\
& TLM & 8.63 & 22.23 & \bestscore{26.20} & 21.66 & 19.68 & 28.60 & 32.76 & 21.37 & 27.58 \\
& \multicolumn{10}{>{\columncolor{gray!10}}c}{\modewocache} \\
& \method{}-2 & 12.68 & 25.66 & 24.89 & 22.10 & 21.33 & 32.01 & \bestscore{33.78} & 21.64 & 29.14 \\
& \method{}-4 & 18.17 & 25.80 & 25.34 & 23.05 & 23.09 & 32.09 & 33.54 & 22.70 & 29.45 \\
& \method{}-8 & \bestscore{19.78} & \secondscore{25.95} & \secondscore{25.67} & \bestscore{23.88} & \bestscore{23.82} & 33.16 & \secondscore{33.78} & \bestscore{24.78} & \bestscore{30.57} \\
& \method{}-16 & 18.91 & 25.00 & 21.73 & \secondscore{23.64} & 22.32 & 32.20 & 33.18 & \secondscore{24.03} & 29.81 \\
& \multicolumn{10}{>{\columncolor{gray!10}}c}{\modecache} \\
& \method{}-2 & 11.03 & 25.73 & 25.56 & 21.86 & 21.05 & 31.51 & 33.63 & 21.41 & 28.85 \\
& \method{}-4 & 17.11 & 25.11 & 25.25 & 22.39 & 22.47 & \secondscore{33.27} & 33.50 & 24.00 & \secondscore{30.26} \\
& \method{}-8 & \secondscore{19.06} & 25.56 & 25.02 & 23.16 & \secondscore{23.20} & \bestscore{33.63} & 33.32 & 23.43 & 30.12 \\
& \method{}-16 & 15.70 & \bestscore{26.42} & 20.54 & 23.03 & 21.42 & 32.56 & 32.86 & 23.66 & 29.69 \\
\midrule
\multirow{14}{*}{\textsc{Qwen-2.5-14B}}
& Base Model & 8.35 & 17.94 & 9.88 & 16.95 & 13.28 & 22.84 & 28.66 & 18.01 & 23.17 \\
& Tent & 12.38 & 24.21 & 12.82 & 21.30 & 17.68 & 32.76 & 30.24 & 19.27 & 27.42 \\
& EATA & 9.67 & 24.83 & 12.65 & 19.69 & 16.71 & 32.53 & 30.63 & 19.36 & 27.50 \\
& TLM & 8.74 & 23.71 & \bestscore{29.17} & 22.01 & 20.91 & 28.95 & 33.22 & 20.32 & 27.50 \\
& \multicolumn{10}{>{\columncolor{gray!10}}c}{\modewocache} \\
& \method{}-2 & 13.44 & 26.25 & 26.69 & 22.68 & 22.26 & 31.52 & \secondscore{33.65} & 22.25 & 29.14 \\
& \method{}-4 & 17.52 & 25.75 & \secondscore{27.06} & 22.13 & \bestscore{23.12} & 31.41 & \bestscore{33.73} & 23.77 & 29.64 \\
& \method{}-8 & \bestscore{19.78} & \bestscore{27.07} & 19.73 & \bestscore{23.00} & 22.39 & 31.20 & 33.38 & \bestscore{27.11} & \bestscore{30.56} \\
& \method{}-16 & 15.73 & 24.81 & 26.57 & 22.63 & 22.43 & 32.53 & 32.98 & \secondscore{24.11} & 29.87 \\
& \multicolumn{10}{>{\columncolor{gray!10}}c}{\modecache} \\
& \method{}-2 & 11.47 & 25.74 & 25.79 & 21.04 & 21.01 & 32.33 & 33.47 & 21.08 & 28.96 \\
& \method{}-4 & 16.97 & 26.00 & 25.27 & 22.32 & \secondscore{22.64} & \bestscore{32.97} & 33.44 & 23.24 & 29.88 \\
& \method{}-8 & 13.38 & 26.50 & 20.69 & \secondscore{22.86} & 20.86 & \secondscore{32.83} & 33.47 & 23.76 & \secondscore{30.02} \\
& \method{}-16 & \secondscore{19.58} & \secondscore{26.58} & 17.21 & 22.25 & 21.41 & 32.77 & 32.49 & 22.91 & 29.39 \\
\bottomrule
\end{tabular}
}
\end{table*}

%% file: tex/appendix_table_bleu.tex
\begin{table*}[t]
\small
\centering
\caption{\small{Detailed results on \texttt{DomainBench} and \texttt{InstructBench}. BLEU scores ($\times 100$ for visibility; higher is better). For each model and dataset, the highest score is \textbf{bold} and the second-highest is \underline{underlined}.}}
\vspace{0.5em}
\label{tab:appendix_table_bleu}
\resizebox{\textwidth}{!}{%
\begin{tabular}{l l *{4}{S} S *{3}{S} S}
\toprule
& & \multicolumn{5}{c}{\textbf{\texttt{DomainBench}}} & \multicolumn{4}{c}{\textbf{\texttt{InstructBench}}} \\
\cmidrule(lr){3-7} \cmidrule(lr){8-11}
\textbf{Model} & \textbf{Method} &
\texttt{Agriculture} & \texttt{GeoSignal} & \texttt{GenMedGPT} & \texttt{Wealth} &
\multicolumn{1}{c}{\textbf{Avg.}} &
\texttt{Dolly} & \texttt{Alpaca-GPT4} & \texttt{InstructWild} &
\multicolumn{1}{c}{\textbf{Avg.}} \\
\midrule
\multirow{14}{*}{\textsc{Llama-3.2-3B}}
& Base Model & 1.00 & 4.71 & 1.65 & 3.11 & 2.62 & 7.64 & 8.68 & 3.35 & 6.55 \\
& Tent & 1.04 & 4.24 & 1.59 & 2.15 & 2.26 & 7.10 & 10.55 & 3.66 & 7.10 \\
& EATA & 1.08 & 6.78 & 2.01 & 2.99 & 3.21 & 6.20 & 8.73 & 2.92 & 5.95 \\
& TLM & 2.24 & 7.96 & \secondscore{5.21} & 5.33 & \secondscore{5.18} & 6.31 & 10.72 & 4.50 & 7.18 \\
& \multicolumn{10}{>{\columncolor{gray!10}}c}{\modewocache} \\
& \method{}-2 & \secondscore{3.32} & \bestscore{8.34} & \bestscore{5.39} & \bestscore{6.42} & \bestscore{5.87} & 8.35 & 12.47 & 7.08 & 9.30 \\
& \method{}-4 & 3.27 & 7.43 & 2.50 & \secondscore{6.39} & 4.90 & 9.52 & \secondscore{12.68} & 7.86 & \secondscore{10.02} \\
& \method{}-8 & 3.15 & 8.11 & 2.74 & 6.34 & 5.08 & 9.64 & 12.23 & 7.38 & 9.75 \\
& \method{}-16 & 3.12 & 4.76 & 2.40 & 5.60 & 3.97 & 6.84 & 10.55 & 5.83 & 7.74 \\
& \multicolumn{10}{>{\columncolor{gray!10}}c}{\modecache} \\
& \method{}-2 & 2.44 & 7.61 & 2.52 & 6.25 & 4.70 & \bestscore{10.37} & 11.96 & 6.22 & 9.51 \\
& \method{}-4 & \bestscore{3.58} & \secondscore{8.12} & 2.51 & 6.36 & 5.14 & 10.00 & \bestscore{12.72} & \bestscore{8.47} & \bestscore{10.40} \\
& \method{}-8 & 2.81 & 7.80 & 3.44 & 6.08 & 5.03 & \secondscore{10.07} & 11.65 & \secondscore{7.88} & 9.86 \\
& \method{}-16 & 2.26 & 7.43 & 2.40 & 5.57 & 4.41 & 9.79 & 10.12 & 5.83 & 8.58 \\
\midrule
\multirow{14}{*}{\textsc{Llama-3.1-8B}}
& Base Model & 1.10 & 4.69 & 1.54 & 3.10 & 2.61 & 9.31 & 8.70 & 3.30 & 7.10 \\
& Tent & 1.37 & 6.45 & 2.16 & 3.78 & 3.44 & 7.12 & 5.39 & 4.37 & 5.63 \\
& EATA & 1.06 & 6.20 & 1.71 & 1.12 & 2.52 & 10.04 & 1.12 & 4.10 & 5.09 \\
& TLM & 2.78 & 8.08 & 6.27 & 6.21 & 5.83 & 9.48 & 10.57 & 4.81 & 8.29 \\
& \multicolumn{10}{>{\columncolor{gray!10}}c}{\modewocache} \\
& \method{}-2 & 2.96 & \bestscore{8.99} & \secondscore{6.98} & \secondscore{6.74} & \secondscore{6.42} & 10.27 & \secondscore{12.87} & 5.55 & 9.56 \\
& \method{}-4 & 3.52 & \secondscore{8.63} & \bestscore{7.57} & 6.65 & \bestscore{6.59} & 10.04 & \bestscore{12.95} & \bestscore{8.51} & \secondscore{10.50} \\
& \method{}-8 & \secondscore{3.59} & 8.27 & 5.48 & \bestscore{7.13} & 6.12 & 9.43 & 12.84 & 7.56 & 9.94 \\
& \method{}-16 & \bestscore{3.63} & 6.45 & 6.17 & 5.75 & 5.50 & 7.12 & 11.04 & \secondscore{8.21} & 8.79 \\
& \multicolumn{10}{>{\columncolor{gray!10}}c}{\modecache} \\
& \method{}-2 & 2.65 & 8.35 & 3.78 & 6.66 & 5.36 & 10.69 & 12.58 & 5.24 & 9.50 \\
& \method{}-4 & 2.44 & 8.20 & 5.57 & 6.46 & 5.67 & \bestscore{12.01} & 12.84 & 8.19 & \bestscore{11.01} \\
& \method{}-8 & 2.71 & 8.05 & 6.93 & 6.53 & 6.05 & \secondscore{10.88} & 12.31 & 6.76 & 9.98 \\
& \method{}-16 & 2.49 & 7.85 & 3.97 & 6.23 & 5.14 & 10.24 & 11.92 & 6.64 & 9.60 \\
\midrule
\multirow{14}{*}{\textsc{Qwen-2.5-7B}}
& Base Model & 1.22 & 4.48 & 1.23 & 3.80 & 2.68 & 6.62 & 10.48 & 3.75 & 6.95 \\
& Tent & 3.54 & 8.01 & 1.79 & 6.17 & 4.88 & 8.57 & 12.92 & 5.49 & 8.99 \\
& EATA & 2.32 & 8.11 & 1.70 & 5.90 & 4.51 & 8.74 & 10.85 & 6.27 & 8.62 \\
& TLM & 1.47 & 7.25 & \bestscore{9.98} & 6.29 & 6.25 & 10.70 & 13.14 & 6.07 & 9.97 \\
& \multicolumn{10}{>{\columncolor{gray!10}}c}{\modewocache} \\
& \method{}-2 & 2.59 & 8.66 & 9.24 & 6.47 & 6.74 & 11.62 & \bestscore{14.39} & 6.21 & 10.74 \\
& \method{}-4 & 3.66 & 8.68 & \secondscore{9.69} & 6.84 & 7.22 & 12.42 & 13.77 & 6.84 & 11.01 \\
& \method{}-8 & \bestscore{4.35} & 8.82 & 9.58 & \bestscore{7.27} & \bestscore{7.51} & 12.33 & \secondscore{14.35} & \bestscore{7.75} & \bestscore{11.48} \\
& \method{}-16 & 3.94 & 8.29 & 7.43 & \secondscore{7.27} & 6.73 & 12.16 & 13.59 & 7.21 & 10.98 \\
& \multicolumn{10}{>{\columncolor{gray!10}}c}{\modecache} \\
& \method{}-2 & 2.12 & \bestscore{8.88} & 9.68 & 6.37 & 6.76 & 11.70 & 14.32 & 6.12 & 10.71 \\
& \method{}-4 & 3.39 & 8.53 & 9.34 & 6.56 & 6.96 & \secondscore{12.92} & 13.92 & 7.46 & \secondscore{11.43} \\
& \method{}-8 & \secondscore{4.01} & \secondscore{8.86} & 9.28 & 6.95 & \secondscore{7.27} & \bestscore{13.04} & 13.84 & 7.35 & 11.41 \\
& \method{}-16 & 3.09 & 8.51 & 6.99 & 6.95 & 6.38 & 12.20 & 14.27 & \secondscore{7.47} & 11.31 \\
\midrule
\multirow{14}{*}{\textsc{Qwen-2.5-14B}}
& Base Model & 1.46 & 4.99 & 1.49 & 3.92 & 2.97 & 6.77 & 11.37 & 3.99 & 7.38 \\
& Tent & 2.27 & 7.67 & 1.81 & 6.22 & 4.49 & 11.85 & 12.11 & 4.47 & 9.47 \\
& EATA & 1.63 & 7.24 & 1.95 & 5.45 & 4.07 & 11.71 & 13.07 & 5.08 & 9.95 \\
& TLM & 1.46 & 8.10 & \bestscore{11.10} & 6.44 & 6.77 & 10.63 & 14.12 & 5.50 & 10.08 \\
& \multicolumn{10}{>{\columncolor{gray!10}}c}{\modewocache} \\
& \method{}-2 & 2.58 & \secondscore{9.41} & \secondscore{9.96} & 6.71 & \secondscore{7.16} & 12.07 & \bestscore{14.95} & 6.55 & 11.19 \\
& \method{}-4 & 3.40 & 9.39 & 9.86 & 6.51 & \bestscore{7.29} & 11.53 & \secondscore{14.89} & 7.26 & 11.23 \\
& \method{}-8 & \secondscore{4.00} & 9.04 & 6.07 & \bestscore{6.96} & 6.52 & 11.87 & 14.63 & \bestscore{8.78} & \bestscore{11.76} \\
& \method{}-16 & 3.12 & 8.25 & 9.58 & 6.79 & 6.94 & 11.71 & 13.86 & \secondscore{7.48} & 11.02 \\
& \multicolumn{10}{>{\columncolor{gray!10}}c}{\modecache} \\
& \method{}-2 & 2.12 & 8.71 & 9.37 & 6.03 & 6.56 & \bestscore{12.55} & 14.81 & 5.82 & 11.06 \\
& \method{}-4 & 3.50 & 8.79 & 8.91 & 6.54 & 6.94 & \secondscore{12.28} & 14.83 & 7.08 & \secondscore{11.40} \\
& \method{}-8 & 2.41 & 9.22 & 6.03 & \secondscore{6.93} & 6.15 & 12.18 & 14.34 & 7.33 & 11.28 \\
& \method{}-16 & \bestscore{4.24} & \bestscore{9.62} & 4.23 & 6.46 & 6.14 & 12.13 & 14.05 & 7.11 & 11.10 \\
\bottomrule
\end{tabular}
}
\end{table*}